\documentclass[journal]{IEEEtran}
\usepackage{cite}
\usepackage{amsmath,amssymb,amsfonts}
\usepackage{algorithmic}
\usepackage{graphicx}
\usepackage{caption}
\usepackage[table]{xcolor}
\usepackage{textcomp}
\usepackage{wrapfig}
\usepackage{float}
\usepackage{stfloats}
\usepackage{placeins}
\usepackage{needspace}
\usepackage{booktabs}
\usepackage{multirow}
\usepackage{array}
\usepackage{colortbl}
\usepackage{tabularx}
\usepackage{url}
\usepackage[normalem]{ulem}
\usepackage{soul}
\usepackage[hidelinks]{hyperref}

\def\BibTeX{{\rm B\kern-.05em{\sc i\kern-.025em b}\kern-.08em
    T\kern-.1667em\lower.7ex\hbox{E}\kern-.125emX}}
\definecolor{abstractbg}{rgb}{1,1,1}
\definecolor{subsectioncolor}{rgb}{0,0,0}
\definecolor{AblationBestBlue}{RGB}{232,244,255}
\definecolor{AblGreen}{RGB}{0,155,85}
\definecolor{AblRed}{RGB}{220,20,20}
\definecolor{AlgGreen}{RGB}{0,165,55}
\definecolor{AlgMagenta}{RGB}{218,0,135}
\definecolor{AlgBlue}{RGB}{0,92,160}
\definecolor{AlgPurple}{RGB}{86,54,140}
\definecolor{AlgGray}{RGB}{45,45,45}
\definecolor{FeatYellow}{RGB}{255,195,0}
\definecolor{FeatOrange}{RGB}{255,100,25}
\newcommand{\cmark}{\textcolor{AblGreen}{\(\checkmark\)}}
\newcommand{\xmark}{\textcolor{AblRed}{\(\times\)}}
\newcommand{\figpanel}[1]{\textcolor{subsectioncolor}{(#1)}}
\newcolumntype{C}{>{\centering\arraybackslash}X}
\newcounter{ssealgorithm}
\newcommand{\algtitle}[1]{\refstepcounter{ssealgorithm}\noindent\textbf{Algorithm~\thessealgorithm}~#1}
\newcommand{\algcomment}[1]{\textcolor{AlgGreen}{\# #1}}
\newcommand{\algkw}[1]{\textcolor{AlgMagenta}{#1}}

\newcommand{\algop}[1]{\textcolor{AlgPurple}{#1}}
\newcommand{\algstage}[1]{\textcolor{AlgBlue}{\(\triangleright\) #1}}
\newcommand{\algindent}{\hspace*{1.35em}}
\newif\ifmarkedchanges
\ifdefined\ForceMarkedChanges
  \markedchangestrue
\else
  \markedchangesfalse
\fi
\newcommand{\rev}[1]{%
  \ifmarkedchanges
    {\color{blue}\uwave{#1}}\normalcolor
  \else
    #1%
  \fi}
\newcommand{\revdel}[1]{%
  \ifmarkedchanges
    {\color{red}\st{#1}}\normalcolor
  \fi}
\newcommand{\revsep}{%
  \ifmarkedchanges\hspace{0.25em}\fi}
\let\sseoriginalcite\cite
\newcommand{\ssebluecite}[1]{%
  \ifmarkedchanges
    \mbox{\textcolor{blue}{\sseoriginalcite{#1}}}%
  \else
    \sseoriginalcite{#1}%
  \fi}
\newcommand{\addedcite}[1]{%
  \ssebluecite{#1}}
\newcommand{\deletedcite}[1]{%
  \ssebluecite{#1}}
\soulregister\deletedcite7
\soulregister\addedcite7
\newcommand{\shiftedcite}[1]{%
  \ssebluecite{#1}}
\soulregister\shiftedcite7
\soulregister\ref7
\soulregister\eqref7
\newcommand{\addedsubsection}[2]{%
  \ifmarkedchanges
    \begingroup
    \renewcommand{\thesubsectiondis}{%
      \textcolor{blue}{\uwave{#1}}.}%
    \subsection{\texorpdfstring{\rev{#2}}{#2}}%
    \endgroup
  \else
    \subsection{#2}%
  \fi}
\newcommand{\renumberedsubsection}[3]{%
  \ifmarkedchanges
    \begingroup
    \renewcommand{\thesubsectiondis}{%
      \textcolor{red}{\sout{#1}}\hspace{0.2em}%
      \textcolor{blue}{\uwave{#2}}.}%
    \subsection{#3}%
    \endgroup
  \else
    \subsection{#3}%
  \fi}
\newcommand{\colorsafesubsection}[2]{%
  \begingroup
  \renewcommand{\thesubsectiondis}{\textcolor{subsectioncolor}{\Alph{subsection}.}}%
  \subsection{\texorpdfstring{\textcolor{subsectioncolor}{#1}}{#1}}%
  \label{#2}%
  \endgroup}

\begin{document}
\title{S$^3$E-Det: A Strict Spatial--Spectral Equivariant Detector for Multispectral Object Detection}
\author{Peng Zhang, Tingfa Xu\textsuperscript{*}, Jinrui Sui, Yuyang Tian, and Jianan Li\textsuperscript{*}
\thanks{Peng Zhang, Jinrui Sui, and Yuyang Tian are with Beijing Institute of Technology, Beijing 100081, China. Tingfa Xu and Jianan Li are with Beijing Institute of Technology, Beijing 100081, China (e-mail: 3120240593@bit.edu.cn; sui2324129420@163.com; 3120250621@bit.edu.cn; ciom\_xtf1@bit.edu.cn; lijianan@bit.edu.cn).}
\thanks{Tingfa Xu is also with Big Data and Artificial Intelligence Laboratory, Beijing Institute of Technology Chongqing Innovation Center, Chongqing 401135, China.}
\thanks{\textsuperscript{*}Corresponding authors: Tingfa Xu and Jianan Li.}}

\IEEEtitleabstractindextext{%
\begin{abstract}
Multispectral object detection requires detectors to exploit material-dependent spectral responses while maintaining consistent localization under geometric variations.\spacefactor=1000\relax\ %
\revdel{}\revsep%
\revdel{}\revsep%
\rev{\mbox{Existing} MOD detectors learn discriminative spectral representations, yet lack a unified mechanism that preserves spectral order, adapts spectral--context fusion across pyramid levels, and regulates geometric responses under rotation.}
To address these \revdel{}\revsep\rev{limitations}, we propose S$^3$E-Det, a strict spatial--spectral equivariant detector \revdel{}\revsep\rev{that couples ordered spectral modeling with transformation-consistent prediction}.\spacefactor=1000\relax\ %
\revdel{}\revsep%
\revdel{}\revsep%
\rev{Spectral Order-aware Sampling Calibration (SOSC) imposes local order constraints on input-derived spectral features through first-order variation, while Common-Differential Equivariant Fusion (CDEF) adapts spectral--context aggregation across pyramid levels.}\spacefactor=1000\relax\ %
\revdel{}\revsep%
\rev{A compact Geometric Alignment Head (GAH) reconciles invariant and equivariant output properties within a unified prediction stage.}\spacefactor=1000\relax\ %
\revdel{}\revsep%
\revdel{}\revsep%
\rev{Across MODA, HOD3K, and DroneVehicle, S$^3$E-Det combines high detection accuracy with a compact parameter footprint and reduced rotation-induced variation.}\spacefactor=1000\relax\ %
Code is available at \textcolor[rgb]{0.07843,0.47843,1}{\url{https://github.com/Riiluo/S3E-Det}}.
\end{abstract}
\begin{IEEEkeywords}
remote sensing, multispectral object detection, rotation-equivariant neural networks, spectral order modeling, spatial-spectral feature learning
\end{IEEEkeywords}}

\maketitle
\IEEEdisplaynontitleabstractindextext
\IEEEpeerreviewmaketitle

\section{Introduction}
\label{sec:introduction}

\begin{figure*}[t]
\centering
\includegraphics[width=\textwidth]{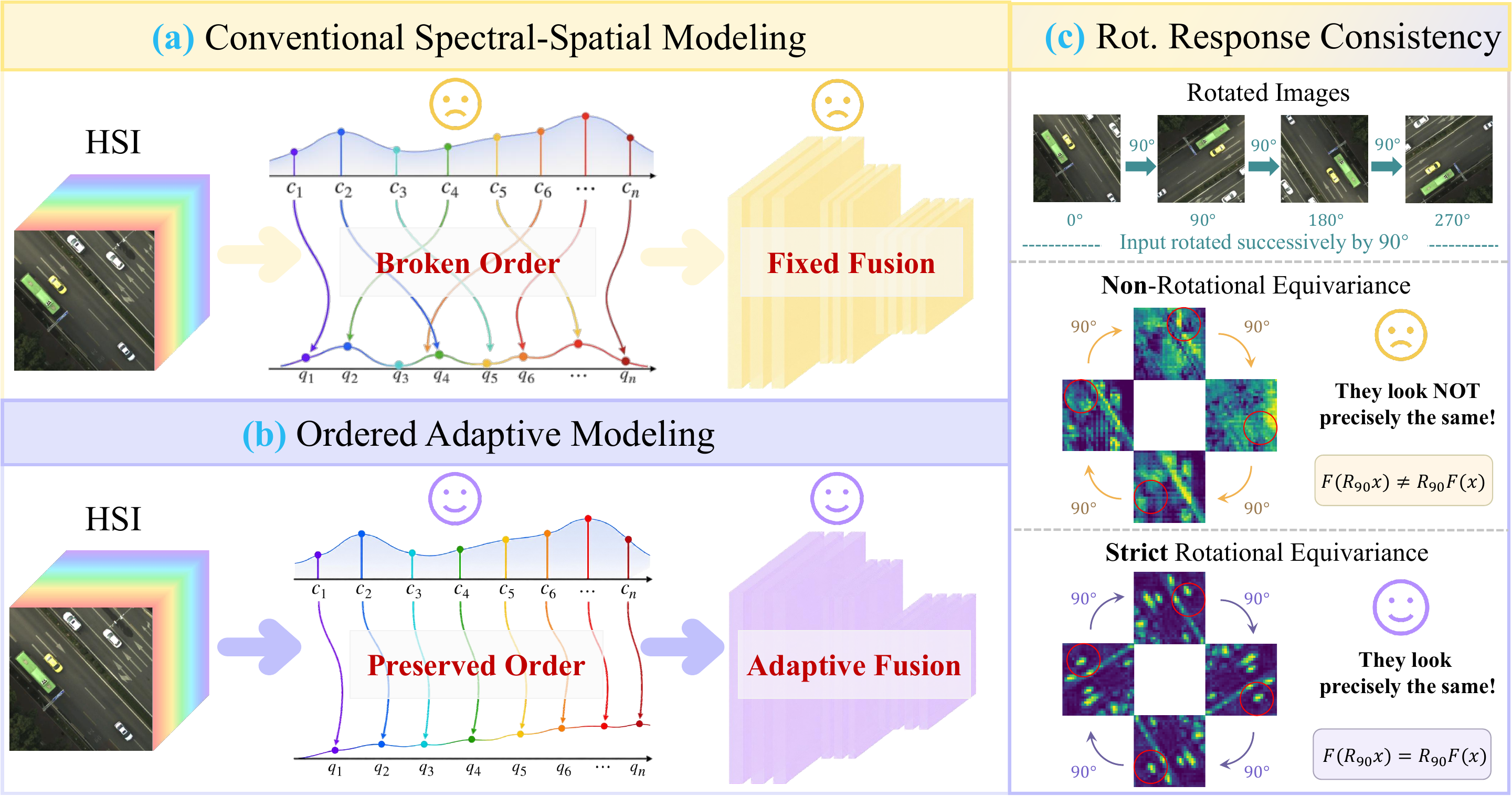}
\caption{\figpanel{a} Conventional spatial--spectral modeling \revdel{}\revsep\rev{does not preserve} the ordered spectral structure and use\rev{s} fixed feature fusion. \figpanel{b} Ordered adaptive modeling preserves spectral order and adapts fusion across feature levels. \figpanel{c} Strict rotational response consistency makes feature responses transform predictably under exact quarter-turn rotations.}
\label{fig:motivation}
\end{figure*}

Multispectral images (MSIs) capture scene reflectance across multiple spectral bands, revealing material-dependent spectral signatures beyond RGB imagery~\shiftedcite{vali2020deep}. Multispectral object detection (MOD) is a fundamental task for remote sensing scene understanding, with applications in traffic monitoring, urban management, and public safety~\shiftedcite{dronevehicle,han2025moda}.\spacefactor=1000\relax\space%
\revdel{}\revsep%
\revdel{}\revsep%
\rev{Existing MOD methods reduce spectral redundancy through band selection and principal-component analysis~\addedcite{he2023s2adet}, or model cross-band dependencies through modulation, masking, and attention~\addedcite{han2025moda}.}\spacefactor=1000\relax\space%
\revdel{}\revsep%
\revdel{}\revsep%
\rev{These operations improve feature discrimination but remain centered on channel selection and transformation, without explicitly modeling the local ordering of spectral bands.}\spacefactor=1000\relax\space%
\revdel{}\revsep%
\rev{Their fusion designs likewise do not condition the balance between local differential spectral evidence and high-level object context on pyramid level.}\spacefactor=1000\relax\space%
\rev{Thus, MOD requires spectral-order preservation and scale-adaptive fusion (\mbox{Figure}~\ref{fig:motivation}(a)).}

\setlength{\parskip}{0pt}
\revdel{}\revsep
\rev{These spectral limitations are compounded in oriented MOD because detection variables do not all obey the same transformation law under input rotation.}\spacefactor=1000\relax\ %
\revdel{}\revsep
\rev{Existing MOD pipelines rely on increased model capacity or rotation augmentation to learn rotational behavior implicitly, without explicitly constraining how spectral features or individual detector outputs should transform under rotation.}\spacefactor=1000\relax\ %
\revdel{}\revsep
\revdel{}\revsep
\revdel{}\revsep
\revdel{}\revsep
\revdel{}\revsep
\rev{Spatial equivariance alone is insufficient. Equivariant MOD must preserve local band order, adapt spectral--context fusion across scales, and assign rotation laws to heterogeneous outputs. This formulation reconciles invariant class and canonical-geometry readouts with equivariant orientation prediction (Figures~\ref{fig:motivation}(b) and~\ref{fig:motivation}(c)).}

\rev{Three coupled limitations remain.} \rev{First, band reduction, reweighting, and unconstrained spatial--spectral mixing do not preserve sensor-defined band order or its derived latent topology.} \rev{Second, level-agnostic fusion cannot adjust local differential spectral evidence and high-level object context as resolution and semantic depth change.} \rev{Third, current MOD pipelines lack variable-level rotation constraints that reconcile invariant and equivariant outputs within one prediction framework under response-specific transformation laws.}
\setlength{\parskip}{0ex plus 0.2ex minus 0.1ex}

\revdel{} \revdel{} \revdel{} \revdel{}
\setlength{\parskip}{0pt}
{\clubpenalty=10000
\revdel{} \rev{To address these limitations,} we propose S$^3$E-Det, a strict spatial--spectral equivariant detector for MOD. This design ties spectrally aware sampling and multi-scale aggregation to the same equivariant feature space. Specifically, Spectral Order-aware Sampling Calibration (SOSC) preserves the physical order of the input bands and constrains the input-derived latent channel topology within each block through local order-aware sampling and first-order variation. The resulting representations are \revdel{}\revsep propagated across scales through Common-Differential Equivariant Fusion (CDEF), which \revdel{}\revsep\rev{balances spectral evidence and object context according to pyramid resolution and semantic depth.}\par}

{\clubpenalty=10000
\revdel{} \revdel{} \revdel{}\revsep\rev{To satisfy these response-specific geometric requirements,} S$^3$E-Det adopts a lightweight Geometric Alignment Head (GAH), which decodes detection variables from shared group-indexed feature representations through response-specific readouts. GAH couples invariant category estimation with equivariant localization in a single prediction stage. The three modules preserve spectral structure and equivariant propagation through compact, response-aligned outputs.\par}

\begin{figure}[!b]
\centering
\includegraphics[width=\columnwidth]{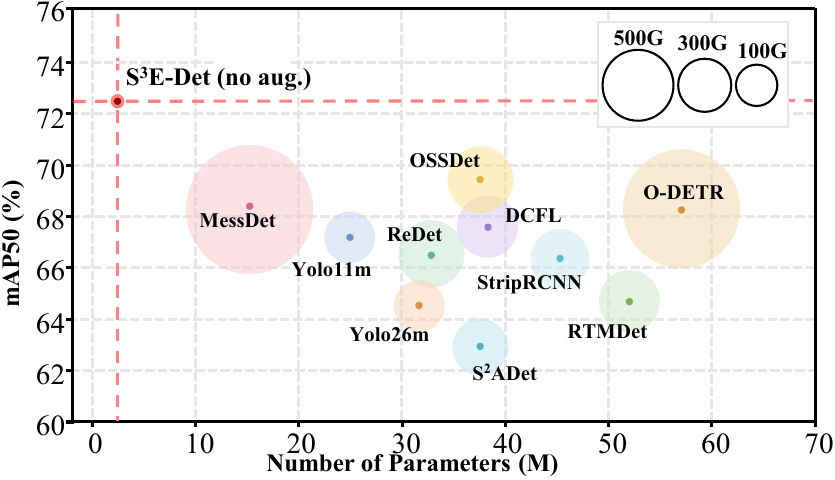}
\caption{mAP$_{50}$ vs. Params on MODA; bubble area: FLOPs.}
\label{fig:moda_efficiency}
\end{figure}

\revdel{}\revsep\rev{Experiments on MODA, HOD3K, and DroneVehicle evaluate the same detector across complementary multispectral configurations and scene conditions.} \revdel{}\revsep\rev{Across these settings, S$^3$E-Det combines strong accuracy, stable rotation response, and a compact parameter footprint.} \revdel{}\revsep\rev{\mbox{Figure~\ref{fig:moda_efficiency}} summarizes the mAP$_{50}$--parameter trade-off on MODA.} \revdel{}\revsep\rev{The main contributions are}:

\setlength{\parskip}{0ex plus 0.2ex minus 0.1ex}
\begin{itemize}
    \item We propose S$^3$E-Det, a strict spatial--spectral equivariant detector \revdel{}\revsep\rev{that couples ordered spectral structure, group-indexed multi-scale propagation, and geometry-aligned prediction in a unified framework}.
    \item We introduce Spectral Order-aware Sampling Calibration (SOSC), which \revdel{}\revsep\rev{combines block-local latent spectral cells with first-order variation under order-preserving calibration during encoding}.
    \item We develop Common-Differential Equivariant Fusion (CDEF)\revdel{}\revsep\rev{ for scale-adaptive common--differential updates}, paired with a compact Geometric Alignment Head (GAH)\revdel{}\revsep\rev{ for response-specific invariant and equivariant readouts}.
\end{itemize}

\FloatBarrier

\section{Related Work}

\addedsubsection{A}{Spectral Image Analysis}

\rev{Multispectral and hyperspectral imaging capture wavelength-resolved, material-dependent signatures beyond conventional RGB imagery~\addedcite{hong2026hyperspectral}.} \rev{Spectral image analysis spans classification~\addedcite{ma2025lightweight}, object detection~\addedcite{he2023s2adet}, reconstruction~\addedcite{cai2022mst}, and hyperspectral--multispectral fusion~\addedcite{li2026im2state}.} \rev{Recent work also targets transferable pretraining; SpectralGPT learns generic spatial--\allowbreak spectral representations across downstream tasks~\addedcite{hong2024spectralgpt}.} \rev{Multispectral oriented detection adds geometry-aligned instance perception to this landscape, requiring discriminative spectral--spatial representations and transformation-consistent localization.}

\renumberedsubsection{A}{B}{Multispectral Object Detection}

\revdel{} \rev{Current multispectral detectors improve spectral selection and aggregation but leave three coupled limitations.} \revdel{} \rev{First, S$^2$ADet selects bands and applies principal-component analysis before two-stream spectral--spatial aggregation~\addedcite{he2023s2adet}, whereas OSSDet modulates, masks, and attends across spectral features~\addedcite{han2025moda}.} \revdel{} \rev{These strategies improve discrimination but treat the spectral axis mainly as evidence to select, reweight, or compress without constraining local order in input-derived spectral features; SOSC instead calibrates block-local latent spectral cells.} \revdel{} \rev{Second, their two-stream and attention-based fusion schemes aggregate complementary cues without conditioning common object context and differential spectral evidence on pyramid level; CDEF performs this adaptation within a group-indexed pyramid.} \rev{Third, these published MOD designs do not assign distinct transformation laws to category scores, canonical box distances, and orientation; GAH uses invariant readouts for the first two and an equivariant orientation response.} \revdel{} \rev{S$^3$E-Det couples spectral order, scale adaptation, and geometric consistency in one detector.}

\renumberedsubsection{B}{C}{Rotation-Equivariant Object Detection}

Rotation-equivariant networks originate from G-CNNs~\shiftedcite{cohen2016gcnn}, which define group convolutions for discrete rotations and reflections. ReDet~\shiftedcite{han2021redet}, MessDet~\shiftedcite{wu2025messdet}, and R2Det~\shiftedcite{wu2025r2det} extend this idea to oriented or generic object detection. \revdel{}\revsep\rev{Rotation equivariance is equally relevant beyond remote sensing, particularly for natural-scene objects exhibiting diverse and unconstrained orientations}~\addedcite{wu2025r2det}. Although these methods improve rotational stability, their prediction stages \revdel{}\revsep\rev{do not enforce response-specific transformation laws}, so \revdel{}\revsep\rev{a shared decoding rule does not preserve the} equivariant feature structure. Our geometric alignment head therefore matches response readouts to the group action at low cost.

\section{Method}

\begin{figure*}[t]
\centering
\includegraphics[width=\textwidth]{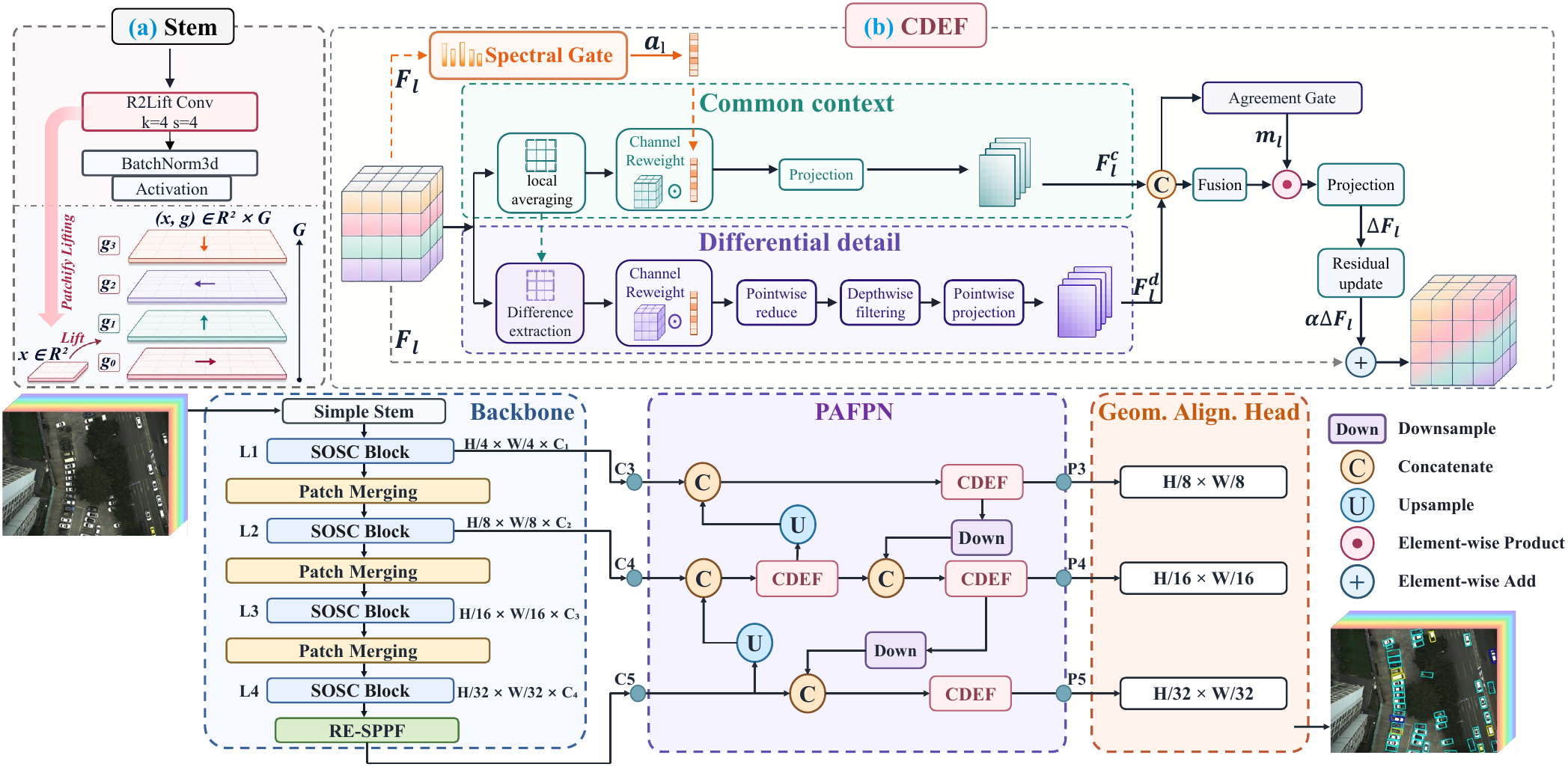}
\caption{Overall network framework of S$^3$E-Det. SOSC blocks perform staged equivariant feature extraction, CDEF conducts multi-scale fusion, and GAH decodes invariant categories and aligned oriented boxes at three scales. \figpanel{a} Stem maps multispectral inputs into group-indexed feature fields; \figpanel{b} CDEF separates common context from differential spectral detail before an agreement-gated residual update.}
\label{fig:network_overview}
\end{figure*}

We present S$^3$E-Det, a spatial--spectral rotation-equivariant detector\revdel{}, as shown in Figure~\ref{fig:network_overview}. \revdel{} \rev{It coordinates representation and transformation laws across the full detector.} \rev{SOSC preserves order during spectral calibration, CDEF adapts common--differential fusion in the group-indexed pyramid, and GAH decodes invariant categories and transformation-consistent boxes.} \revdel{}

\subsection{Preliminaries}
\label{sec:preliminaries}

Let \(C_N=\{r_0,\ldots,r_{N-1}\}\) denote a finite cyclic group of in-plane rotations. The element \(r_k\) rotates the image plane by \(\theta_k=2\pi k/N\). For a multispectral image \(x\in\mathbb{R}^{B\times L\times H\times W}\), with \(L\) ordered spectral bands, we write the spatial action of \(r\in C_N\) as \(\mathcal{T}_{r}x\). This action rotates the image coordinates without reordering the spectral index. \rev{The default implementation uses \(N=4\), yielding \(C_4\); Table~\ref{tab:group_order_ablation} evaluates alternative group orders.}

A mapping \(\Phi\) is rotation equivariant if applying a rotation before feature extraction is equivalent to transforming the extracted feature field:
\begin{equation}
    \Phi(\mathcal{T}_{r}x)=\rho(r)\Phi(x), \quad r\in C_N,
    \label{eq:rotation_equivariance}
\end{equation}
where \(\rho(r)\) is the induced action on the feature representation. The equation states that rotating the input before feature extraction is equivalent to applying the corresponding feature-space action after feature extraction. The familiar notion of invariance is recovered when this output action is the identity, \(h(\mathcal{T}_{r}x)=h(x)\).

After lifting, the feature tensor has shape \(F\in\mathbb{R}^{B\times C\times N\times H\times W}\), where the third axis indexes the \(C_N\) orientation channels. For each spatial position \(u\in\mathbb{Z}^2\) and orientation index \(s\in C_N\), the regular group action is
\begin{equation}
    [\rho(r)F](u,s)=F(r^{-1}u,r^{-1}s), \quad u\in\mathbb{Z}^2,\; s\in C_N.
    \label{eq:regular_group_action}
\end{equation}
Thus, input rotations rotate the spatial coordinates and induce a cyclic permutation of the group-indexed channels.

\subsection{Spectral Order-aware Sampling Calibration}
\label{sec:sosc}

\revdel{} \rev{Sensor-defined order exists at the input, whereas lifting, projection, and patch merging integrate information across bands into latent spectral--spatial features~}\addedcite{han2025moda,he2023s2adet}\rev{. SOSC carries this structure into the hierarchy as a block-local topology rather than a persistent wavelength identity.} \revdel{} \revdel{} \rev{It confines each response to an intra-cell coordinate and adds a first-order variation term, preventing crossing within the current latent block.}

Let \(X\in\mathbb{R}^{B\times C\times G\times H\times W}\) denote a group-indexed feature tensor, where \(C\) indexes learned spectral-feature slots and \(G\) is the number of orientation channels. Within the current SOSC block, we use the unit-spaced canonical coordinates \(s_i=i\), satisfying \(s_1<s_2<\cdots<s_C\), and define slot \(i\)'s cell using neighboring midpoints. These coordinates describe a block-local latent topology rather than a one-to-one correspondence with sensor wavelengths:
\begin{equation}
    \ell_i=
    \begin{cases}
        s_1, & i=1,\\
        \frac{s_{i-1}+s_i}{2}, & i>1,
    \end{cases}
    \quad
    r_i=
    \begin{cases}
        \frac{s_i+s_{i+1}}{2}, & i<C,\\
        s_C, & i=C.
    \end{cases}
    \label{eq:sosc_cell}
\end{equation}
These boundaries are fixed by the ordered sampling geometry rather than learned thresholds, and adjacent cells meet only at shared midpoints. SOSC predicts an intra-cell sampling coordinate using the first-order spectral derivative along the ordered channel centers:
\begin{equation}
    \nabla_s X_i=\frac{X_{i+1}-X_{i-1}}{s_{i+1}-s_{i-1}}, \quad 1<i<C,
    \label{eq:sosc_derivative}
\end{equation}
Boundary channels use one-sided differences. A lightweight spectral encoder \(\phi(\cdot)\), shared across positions and orientations, predicts \(\rho_i\) from \([X_i,\nabla_s X_i]\):
\begin{equation}
    \rho_i=\tanh\left(\phi_i([X,\nabla_s X])\right).
    \label{eq:sosc_rho}
\end{equation}
This coefficient chooses the sampling side and distance within the local cell:
\begin{equation}
    q_i=s_i+[\rho_i]_+(r_i-s_i)-[-\rho_i]_+(s_i-\ell_i),
    \label{eq:sosc_coordinate}
\end{equation}
where \([z]_+=\max(z,0)\). Since \(q_i\) is confined to its own cell and adjacent cells only meet at shared midpoints, the sampled coordinates remain ordered:
\begin{equation}
    q_1\le q_2\le \cdots \le q_C.
    \label{eq:sosc_order}
\end{equation}

\begin{figure}[t]
\centering
\includegraphics[width=\columnwidth]{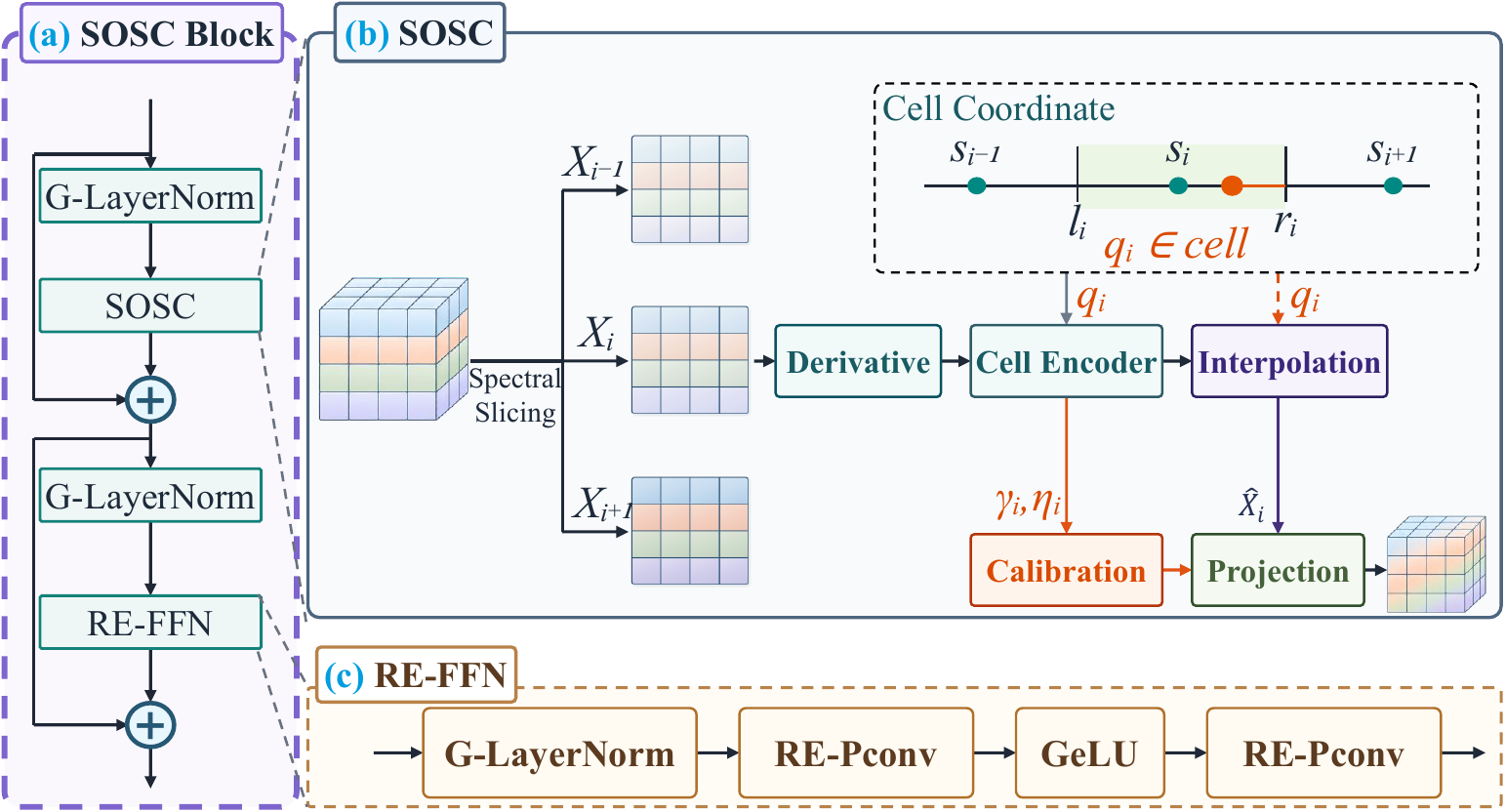}
\caption{Structure of the SOSC block. It calibrates ordered spectral responses through neighboring-support slicing, derivative-aware interpolation, residual update, and RE-FFN.}
\label{fig:sosc_module}
\end{figure}

As shown in Figure~\ref{fig:sosc_module}, an SOSC block acts as a compact spectral-coordinate calibration unit inside each backbone stage. It preserves the stage resolution and returns the calibrated response to the original feature stream.

Thus, SOSC preserves spectral ranking by construction without a hand-tuned displacement bound. It then samples the response at \(q_i\) by linear interpolation:
\begin{equation}
    \hat{X}_i=(1-\omega_i)X_{\lfloor q_i\rfloor}+\omega_i X_{\lceil q_i\rceil},
    \quad
    \omega_i=q_i-\lfloor q_i\rfloor .
    \label{eq:sosc_interpolation}
\end{equation}
The final response keeps \(X_i\) and adds the sampled residual and local spectral slope:
\begin{equation}
    Y_i=X_i+\gamma_i(\hat{X}_i-X_i)+\eta_i\nabla_s X_i,
    \label{eq:sosc_output}
\end{equation}
where \(\gamma_i=\sigma(g_i)\) and \(\eta_i=\tanh(h_i)\) are residual and derivative gates from the same spectral encoder. The residual gate controls the interpolated correction, while the derivative gate injects directional band variation around the original response. Because these predictors are shared across positions and orientations and act only on the spectral axis, SOSC commutes with the regular group action in Eq.~\eqref{eq:regular_group_action}. It refines ordered spectral responses without altering the rotation-equivariant transformation law of the backbone.

\begin{figure*}[!t]
\centering
\includegraphics[width=\textwidth]{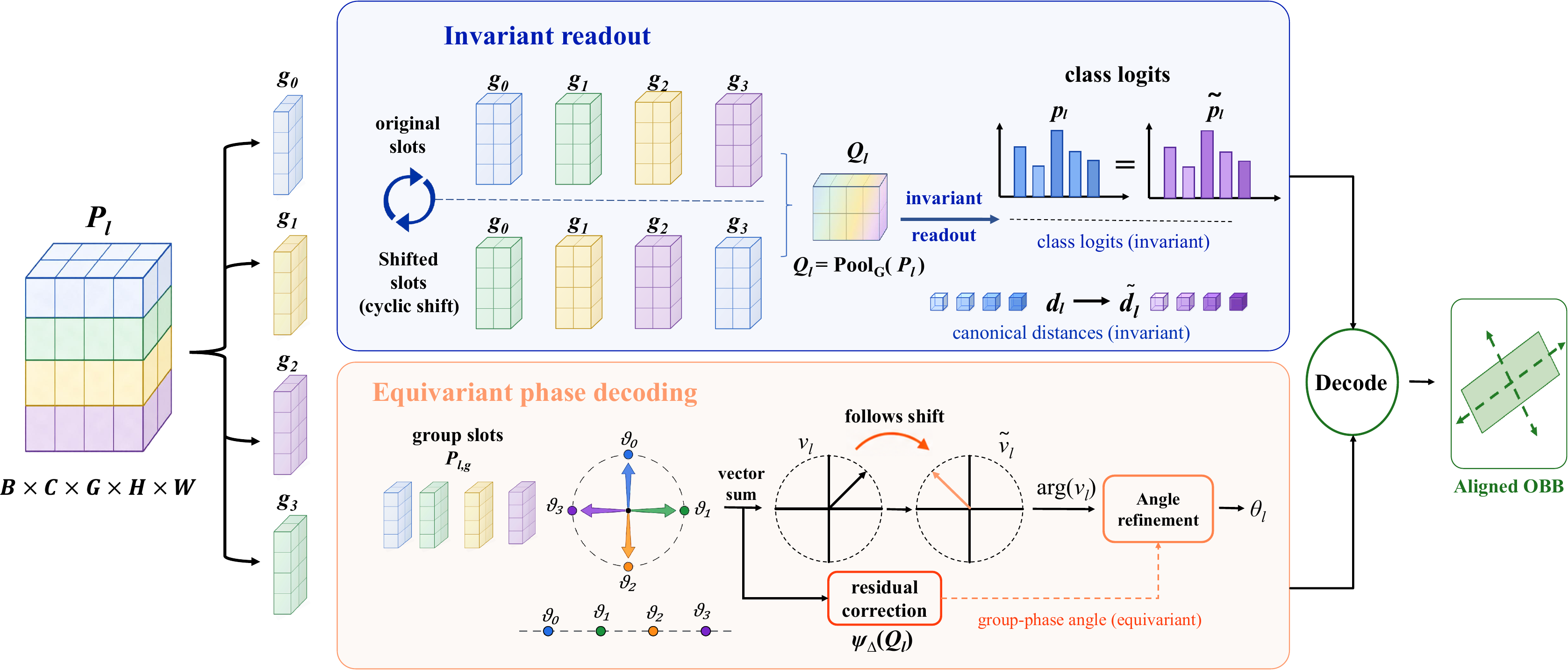}
\caption{Illustration of GAH. The head assigns invariant readouts to category scores and canonical distances, while equivariant group-phase decoding produces the aligned object angle.}
\label{fig:gah_module}
\end{figure*}

\colorsafesubsection{Common-Differential Equivariant Fusion}{sec:cdef}

\revdel{} \rev{Cross-branch fusion strengthens complementarity, but a shared rule cannot balance common object context and differential spectral evidence across scales.} \rev{CDEF makes this balance scale dependent under the same rotation-group action.} \revdel{}\revsep\rev{It} decomposes each group-indexed \revdel{}feature \(F_l\in\mathbb{R}^{B\times C_l\times G\times H_l\times W_l}\) into common and differential evidence.

Figure~\ref{fig:network_overview}(b) gives the CDEF block used in the neck. Rather than replacing the pyramid feature, CDEF estimates a gated equivariant update from common and differential evidence and adds it back to the input feature.

We first estimate a group-invariant spectral channel gate
\begin{equation}
    a_l=\sigma\left(\mathrm{MLP}\left(\mathrm{GAP}_{G,H,W}(F_l)\right)\right),
    \label{eq:cdef_spectral_gate}
\end{equation}
and use it to reweight the feature before decomposition. Let \(\mathcal{A}_k(\cdot)\) denote a local spatial average with kernel size \(k\). The two components are computed as
\begin{subequations}
\label{eq:cdef_common_detail}
\begin{align}
    \mathcal{F}_l^{c}
    &=\psi_c\left(\mathcal{A}_k(F_l)\odot a_l\right), \label{eq:cdef_common}\\
    \mathcal{F}_l^{d}
    &=\psi_d\left(\mathcal{D}_k\left(\psi_r\left((F_l-\mathcal{A}_k(F_l))\odot a_l\right)\right)\right).
    \label{eq:cdef_detail}
\end{align}
\end{subequations}
where \(\psi_c,\psi_r,\psi_d\) are pointwise equivariant projections and \(\mathcal{D}_k\) is an equivariant depthwise spatial operator. Pointwise projections mix channels at each group-indexed position, whereas \(\mathcal{D}_k\) captures within-channel local variation under the same group action. Thus, the common branch aggregates context shared by nearby responses, while the differential branch preserves residual cues for spectrally ambiguous objects.

CDEF fuses the two branches through an agreement gate. We concatenate \(\mathcal{F}_l^{c}\) and \(\mathcal{F}_l^{d}\), project them into a compact equivariant subspace \(Z_l=\psi_f([\mathcal{F}_l^{c},\mathcal{F}_l^{d}])\), and compute
\begin{equation}
    m_l=\sigma(\psi_g(Z_l))\cdot
    \frac{1}{2}\left(1+
    \frac{\left\langle\bar{\mathcal{F}}_l^{c},\bar{\mathcal{F}}_l^{d}\right\rangle}
    {\left\|\bar{\mathcal{F}}_l^{c}\right\|_2\left\|\bar{\mathcal{F}}_l^{d}\right\|_2+\epsilon}\right),
    \label{eq:cdef_agreement}
\end{equation}
where \(\bar{\mathcal{F}}_l^{c}\) and \(\bar{\mathcal{F}}_l^{d}\) denote orientation-averaged responses. The fused output is
\begin{equation}
    F'_l = F_l+\alpha\,\psi_o(Z_l\odot m_l),
    \label{eq:cdef_output}
\end{equation}
where \(\alpha\) is learnable. This output balances gated spectral channels, common object context, and differential local evidence at each pyramid level.

\begin{table*}[t]
\vspace{-4pt}
\centering
\scriptsize
\setlength{\tabcolsep}{1.2pt}
\caption{Comparison with representative methods on MODA. \(\dagger\) indicates methods originally designed for multispectral object detection.}
\vspace{-4pt}
\label{tab:moda_comparison}
\begin{tabular*}{\textwidth}{@{\extracolsep{\fill}}c|l|cccccccc|ccc|cc@{}}
\toprule
 & Methods & Car & Bus & Van & Awn. bike & Truck & Tricycle & Bicycle & Pedestrian & mAP$_{50}\uparrow$ & mAP$_{75}\uparrow$ & mAP$\uparrow$ & \#F$\downarrow$ & \#P$\downarrow$ \\
\midrule
\multirow{4}{*}{\rotatebox[origin=c]{90}{two-stage}} & LSKNet~\shiftedcite{lsknet} & 90.4 & 88.5 & 74.5 & 72.5 & 66.0 & 52.5 & 41.0 & 28.2 & 64.2 & 42.8 & 39.4 & 221.9G & 32.6M \\
 & O-RCNN~\shiftedcite{oriented_rcnn} & 90.5 & 89.7 & 74.6 & 72.5 & 66.6 & 54.7 & 45.1 & 30.4 & 65.5 & 44.0 & 40.9 & 244.7G & 55.3M \\
 & StripRCNN~\shiftedcite{strip_rcnn} & 90.5 & 89.0 & 76.1 & 73.6 & 67.1 & 56.5 & 45.2 & 30.7 & 66.1 & 44.0 & 41.0 & 231.8G & 45.2M \\
 & ReDet~\shiftedcite{han2021redet} & 90.6 & 89.6 & 76.8 & 72.9 & 69.6 & 52.4 & 41.5 & 41.0 & 66.8 & 44.1 & 41.3 & 245.2G & 31.6M \\
\midrule
\multirow{9}{*}{\rotatebox[origin=c]{90}{one-stage}} & R3Det-KLD~\shiftedcite{r3det_kld} & 90.4 & 88.7 & 74.8 & 66.6 & 60.9 & 42.6 & 29.8 & 35.5 & 61.2 & 39.4 & 36.8 & 306.7G & 39.6M \\
 & S$^2$ANet~\shiftedcite{s2anet} & 90.4 & 88.7 & 74.4 & 69.1 & 62.7 & 48.9 & 30.1 & 40.7 & 63.1 & 38.7 & 36.7 & 216.5G & 38.8M \\
 & YOLOv8n~\shiftedcite{yolov8} & 95.8 & 85.2 & 63.7 & 67.4 & 33.6 & 35.9 & 49.6 & 65.1 & 62.0 & 48.0 & 43.3 & \textbf{12.6G} & 3.2M \\
 & YOLOv12m~\shiftedcite{yolov12} & 96.3 & 85.3 & 68.8 & 74.3 & 40.3 & 36.3 & 57.4 & 69.5 & 66.0 & 52.4 & 46.8 & 115.0G & 24.7M \\
 & O-RepPts~\shiftedcite{oriented_reppoints} & 90.5 & 89.2 & 77.7 & 71.2 & 66.2 & 53.1 & 43.0 & 41.1 & 66.5 & 44.1 & 40.9 & 213.6G & 36.9M \\
 & YOLO11m~\shiftedcite{yolo11} & 96.4 & 88.0 & 68.2 & 75.7 & 44.4 & 38.0 & 58.1 & 68.7 & 67.2 & 54.6 & 48.9 & 126.2G & 25.1M \\
 & DCFL~\shiftedcite{dcfl} & 90.6 & 89.7 & 77.5 & 71.3 & 67.2 & 55.5 & 44.2 & 44.0 & 67.5 & 44.8 & 41.5 & 241.0G & 37.0M \\
 & MessDet~\shiftedcite{wu2025messdet} & 90.0 & 89.1 & 75.4 & 71.4 & \textbf{70.2} & 54.9 & 48.4 & 45.5 & 68.1 & 42.4 & 40.1 & 616.0G & 15.1M \\
 & OSSDet$^\dagger$~\shiftedcite{han2025moda} & 90.5 & 89.9 & \textbf{79.2} & 72.7 & 69.7 & \textbf{58.8} & 45.3 & 45.7 & 69.0 & 45.9 & 42.7 & 263.1G & 36.5M \\
\midrule
\multirow{3}{*}{\rotatebox[origin=c]{90}{\shortstack{DETR\\[-1pt]-based}}} & ARS-DETR~\shiftedcite{arsdetr} & 90.5 & 89.2 & 77.2 & 70.6 & 64.8 & 53.6 & 43.2 & 39.7 & 66.1 & 43.4 & 40.5 & 254.0G & 44.2M \\
 & \rev{RHINO}~\addedcite{rhino} & \rev{90.7} & \rev{89.1} & \rev{77.3} & \rev{70.8} & \rev{65.6} & \rev{54.0} & \rev{42.9} & \rev{41.0} & \rev{66.4} & \rev{44.0} & \rev{40.8} & \rev{319.5G} & \rev{50.8M} \\
 & O-DETR~\shiftedcite{ori_detr} & 90.8 & 89.3 & 78.8 & 72.8 & 68.0 & 56.0 & 46.6 & 40.1 & 67.8 & 45.1 & 41.9 & 492.0G & 57.6M \\
\midrule
 & S$^3$E-Det (no aug.) & 96.0 & 94.2 & 71.0 & 79.5 & 61.5 & 49.6 & 59.0 & 70.0 & 72.6 & 59.7 & 54.0 & 35.6G & \textbf{2.4M} \\
 & \rev{S$^3$E-Det (standard aug.)} & \rev{\textbf{97.1}} & \rev{\textbf{94.9}} & \rev{72.5} & \rev{\textbf{80.2}} & \rev{62.9} & \rev{50.3} & \rev{\textbf{60.1}} & \rev{\textbf{71.2}} & \rev{\textbf{73.7}} & \rev{\textbf{60.7}} & \rev{\textbf{54.8}} & \rev{35.6G} & \rev{\textbf{2.4M}} \\
\bottomrule
\end{tabular*}
\vspace{-10pt}
\end{table*}

\subsection{Geometric Alignment Head}
\label{sec:gah}

\revdel{} \rev{Shared detection heads overlook that category evidence and object orientation respond differently to rotation.} \rev{GAH preserves this distinction: }category scores and canonical box distances are invariant to cyclic \revdel{}\revsep\rev{orientation shifts}, \revdel{}\revsep\rev{whereas} angles follow the shift.
For a pyramid feature, an equivariant stem outputs \(P_l\in\mathbb{R}^{B\times C_l\times G\times H_l\times W_l}\). Let \(Q_l=\mathrm{Pool}_G(P_l)\).
\revdel{} \revdel{}
\rev{As illustrated in Figure~\ref{fig:gah_module}, separate readouts preserve the transformation structure established by the backbone and neck.}

\begin{figure}[!t]
\centering
\begin{minipage}{0.98\columnwidth}
\hrule height 0.6pt
\vspace{2pt}
\algtitle{Geometrically aligned prediction}
\label{alg:gah}
\vspace{2pt}
\hrule height 0.35pt
\vspace{3pt}
{\scriptsize\ttfamily\color{AlgGray}
\begin{tabular}{@{}l@{}}
\algcomment{Input: pyramid features \(\{F_l\}_{l=3}^{5}\), group basis \(\{\vartheta_g\}\)}\\
\algcomment{Output: class logits \(\{p_l\}\) and aligned boxes \(\{B_l\}\)}\\
\algkw{for} \(l \in \{3,4,5\}\) \algkw{do}\\
\algindent \(P_l\) = \algop{EquivariantStem}(\(F_l\)), \quad
    \(Q_l\) = \algop{GroupPool}(\(P_l\))\\
\algindent \algstage{Invariant readout}\\
\algindent\algindent \(p_l\) = \algop{ClassReadout}(\(Q_l\)), \quad
    \(d_l\) = \algop{DistanceReadout}(\(Q_l\))\\
\algindent \algstage{Equivariant phase decoding}\\
\algindent\algindent \(\pi_l\) = \algop{OrientationSoftmax}(\(P_l\))\\
\algindent\algindent \(v_l\) = \algop{VectorSum}(\(\pi_l,\{\vartheta_g\}\))\\
\algindent\algindent \(\theta_l\) = \algop{Wrap}\((\arg(v_l)+\algop{AngleResidual}(Q_l))\)\\
\algindent \algstage{Aligned box decoding}\\
\algindent\algindent \(B_l\) = \algop{DecodeOBB}(\(d_l,\theta_l\))\\
\algkw{end for}\\
\algkw{return} \(\{p_l,B_l\}_{l=3}^{5}\)
\end{tabular}}
\vspace{3pt}
\hrule height 0.6pt
\end{minipage}
\end{figure}

Algorithm~\ref{alg:gah} summarizes response-specific decoding across the detection levels. The invariant readout predicts category logits and canonical box distances from the pooled descriptor, whereas the equivariant phase branch converts group responses into aligned orientations before rotated-box decoding.

Classification and canonical box distances are read as
\begin{equation}
    p_l=\psi_{\mathrm{cls}}\left(Q_l\right), \quad
    d_l=\psi_{\mathrm{box}}\left(Q_l\right),
    \label{eq:gah_invariant_readout}
\end{equation}
where \(p_l\) denotes class logits and \(d_l\) denotes box distances in the object canonical frame.

Orientation is decoded from the group phase rather than a free scalar. Let \(\vartheta_g=2\pi g/G\) be the basis angle of channel \(g\). The angle branch forms a phase vector:
\begin{subequations}\label{eq:gah_phase_vector}
\begin{align}
    \pi_{l,g}&=\operatorname{softmax}_{g}\left(\psi_{\theta}(P_{l,g})\right),
    \label{eq:gah_phase_prob}\\
    v_l&=\sum_{g=0}^{G-1}\pi_{l,g}\left(\cos\vartheta_g,\sin\vartheta_g\right).
    \label{eq:gah_phase_sum}
\end{align}
\end{subequations}
A bounded residual refines the group phase:
\begin{equation}
    \theta_l=\operatorname{wrap}\left(\arg(v_l)+
    \frac{\pi}{G}\tanh\left(\psi_{\Delta}(Q_l)\right)\right).
    \label{eq:gah_angle}
\end{equation}
The canonical distances \(d_l\) and aligned angle \(\theta_l\) are then passed to the standard rotated-box decoder. Following the regular group action in Eq.~\eqref{eq:regular_group_action}, GAH assigns different readouts to different response laws: quantities read from \(Q_l\) are invariant, whereas the phase vector follows the cyclic shift of orientation channels. This keeps category and canonical geometry fixed while orientation transforms with the input.

\section{Experiments}

\subsection{Experimental Setup}
\label{sec:experimental_setup}

\noindent\textbf{Datasets and Evaluation Metrics.}
\revdel{} \rev{The three benchmarks were selected for task relevance and complementary coverage rather than scale alone.} \rev{MODA~\addedcite{han2025moda} is the primary benchmark because its eight bands span approximately 395--950~nm and its large set of oriented instances includes small objects, low illumination, and cluttered urban backgrounds; it therefore supports joint evaluation of ordered spectral modeling, multi-scale fusion, and rotation-consistent localization, as well as controlled ablation and rotation analyses.} \rev{HOD3K~\addedcite{he2023s2adet} increases the input to 16 bands over 470--620~nm and includes people, cars, and bikes across diverse real-world scenes, testing whether the same ordered spectral modeling and pyramid fusion remain effective when the band count and category composition change.} \rev{Its natural-scene imagery also provides a complementary setting for assessing the applicability of equivariant modeling beyond aerial scenes.} \revdel{} \rev{DroneVehicle~\addedcite{dronevehicle} provides paired visible--infrared UAV imagery from day to night across urban roads, residential areas, and parking lots, with dense oriented annotations for five vehicle classes; it evaluates the applicability of the proposed method to paired visible--infrared inputs under varied illumination and scene conditions.} \rev{Together, these real-world datasets provide complementary coverage across spectral configurations and practical MOD scenarios.} We report mAP$_{50}$, mAP$_{75}$, mAP, parameters, and FLOPs to compare accuracy, localization quality, model size, and computation.

\noindent\textbf{Implementation Details.}
Experiments are \revdel{}\rev{conducted} on two NVIDIA RTX 3090 GPUs. Unless otherwise specified, the two S$^3$E-Det variants are trained from scratch with AdamW, an initial learning rate of 0.01, a final learning rate of 0.0001, and a weight decay of 0.0005. Input sizes, batch sizes, and schedules are dataset-specific. \rev{Baseline values are taken from the cited studies under their published benchmark protocols and were not uniformly retrained in our codebase.} \revdel{} \rev{The two S$^3$E-Det rows within each dataset share the official split, spectral input, input resolution, optimizer, training schedule, and evaluation implementation; they differ only in augmentation.} \rev{The original setting disables augmentation to isolate the architectural contribution, whereas the standard-augmentation variant uses translation 0.1, scale 0.5, horizontal-flip probability 0.5, and Mosaic probability 1.0; HSV and all remaining augmentation terms are disabled.}

\subsection{Results on the MODA Dataset}

\noindent\textbf{Data and Setups.}
MODA~\shiftedcite{han2025moda} contains 14,041 multispectral images and 330,191 oriented instances from eight categories. Each \(1200\times900\) image records eight bands spanning about 395--950 nm, with low illumination, small objects, and cluttered backgrounds. \revdel{} \rev{We use all eight bands.} \rev{Each $1200\times900$ image is padded to $1216\times928$, so that both spatial dimensions are divisible by the maximum network stride of 32 without rescaling the image content.} \rev{We train for 20 epochs with a batch size of 10.}

\begin{figure*}[!t]
\centering
\includegraphics[width=\textwidth]{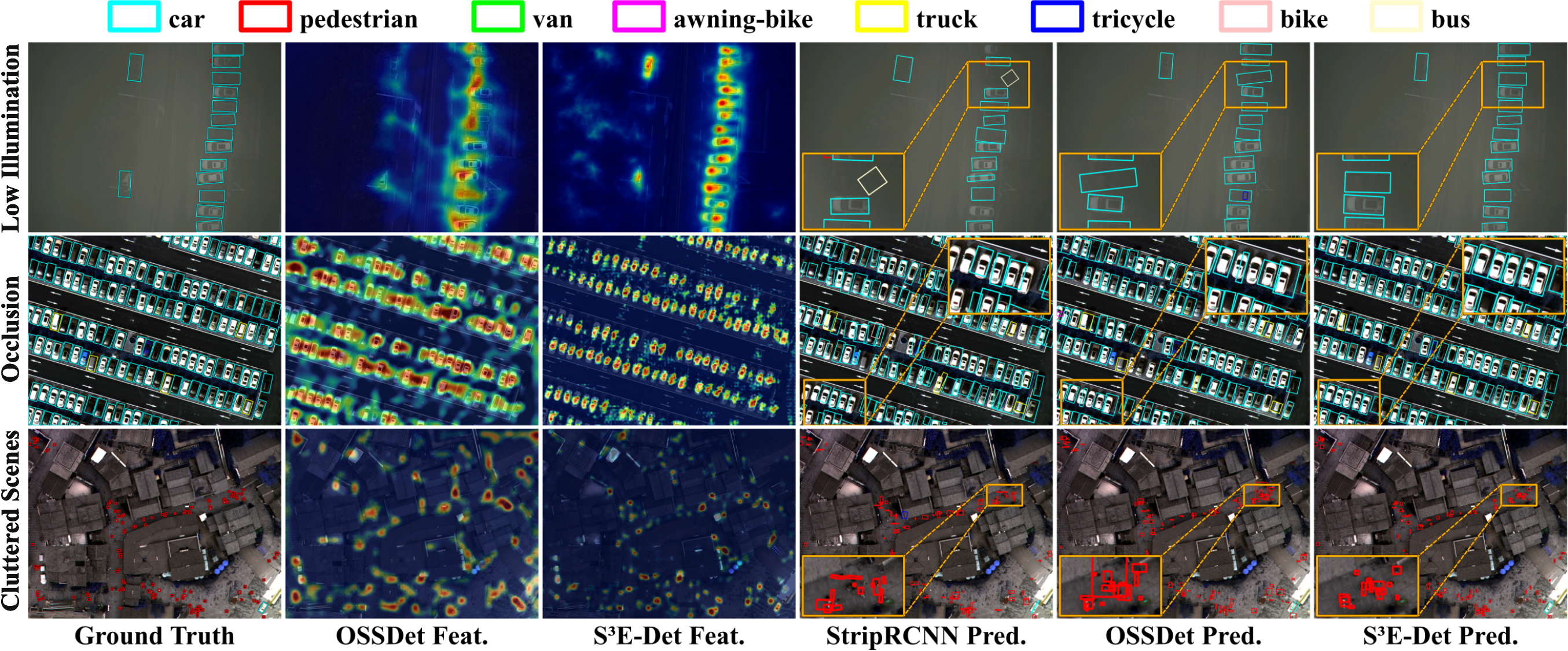}
\caption{Visualization comparison of detection results and feature maps obtained by different methods on MODA.}
\label{fig:moda_qualitative}
\end{figure*}

\noindent\textbf{Quantitative Analysis.}
\revdel{} \rev{Table~\ref{tab:moda_comparison} compares S$^3$E-Det with published general and MOD-specific detector results.} \revdel{} \rev{The two S$^3$E-Det rows provide an internally controlled augmentation comparison, while the cited baseline rows retain their published training protocols.} \revdel{}\revsep\rev{With standard augmentation, S$^3$E-Det gives} the best \revdel{}mAP$_{50}$ and mAP$_{75}$ with the fewest parameters\revdel{}. \revdel{} \rev{Against OSSDet, the absolute differences are 4.7\%, 14.8\%, and 12.1\% in mAP$_{50}$, mAP$_{75}$, and mAP, with 93.3\% fewer parameters and 86.5\% fewer FLOPs.} \revdel{} \revdel{} \rev{S$^3$E-Det also gives the best bus, awning-bike, bicycle, and pedestrian results on MODA.} \rev{Together, the aggregate and class-wise comparisons show that the localization gain extends across multiple categories while retaining a compact model footprint.}

\noindent\textbf{\rev{Computational Cost and Resource Profile.}}
\rev{The default $C_4$ model contains 2,445,776 parameters and requires 35.60G FLOPs for one $1216\times928$ MODA input.} \rev{Table~\ref{tab:eq_pipeline} decomposes these totals by checkpoint and aligns them with conformance residuals without combining the two.} \rev{Table~\ref{tab:inference_efficiency} summarizes parameter count, FLOPs, serialized size, measured peak GPU memory, and per-epoch training time.} \rev{Peak GPU memory is the maximum CUDA allocation observed during the corresponding training run.} \rev{Relative to YOLOv8n, S$^3$E-Det uses fewer parameters while retaining a compact serialized footprint.} \rev{The comparison therefore distinguishes compact model storage from the higher training memory and epoch time of the current implementation.}

\begin{table}[!t]
\centering
\scriptsize
\setlength{\tabcolsep}{2.2pt}
\caption{\rev{Model size and computational resource profile.}}
\label{tab:inference_efficiency}
{%
\ifmarkedchanges\color{blue}\fi
\begin{tabular*}{\columnwidth}{@{\extracolsep{\fill}}lccccc@{}}
\toprule
Model & Params (M) & FLOPs (G) & Size (MB) & \shortstack{Peak GPU Mem\\(GB)} & \shortstack{Time\\(min/epoch)} \\
\midrule
YOLOv8n & 3.2 & 12.6 & 5.75 & 14 & 10 \\
S$^3$E-Det ($C_4$) & 2.4 & 35.6 & 5.43 & 20 & 12 \\
\bottomrule
\end{tabular*}
\vspace{-4pt}
}
\end{table}

\noindent\textbf{Qualitative Analysis.}
\revdel{}\revsep\rev{Figure~\ref{fig:moda_qualitative} shows more target-centered responses and fewer false alarms or misses for small, weakly illuminated, and cluttered objects. The response patterns align with the intended roles of SOSC and CDEF in spectral calibration and scale-aware fusion.} \rev{Crowded scenes remain challenging, with occasional failures to separate adjacent vehicles and detect small pedestrians.}

\begin{figure}[!t]
\centering
\includegraphics[width=\columnwidth]{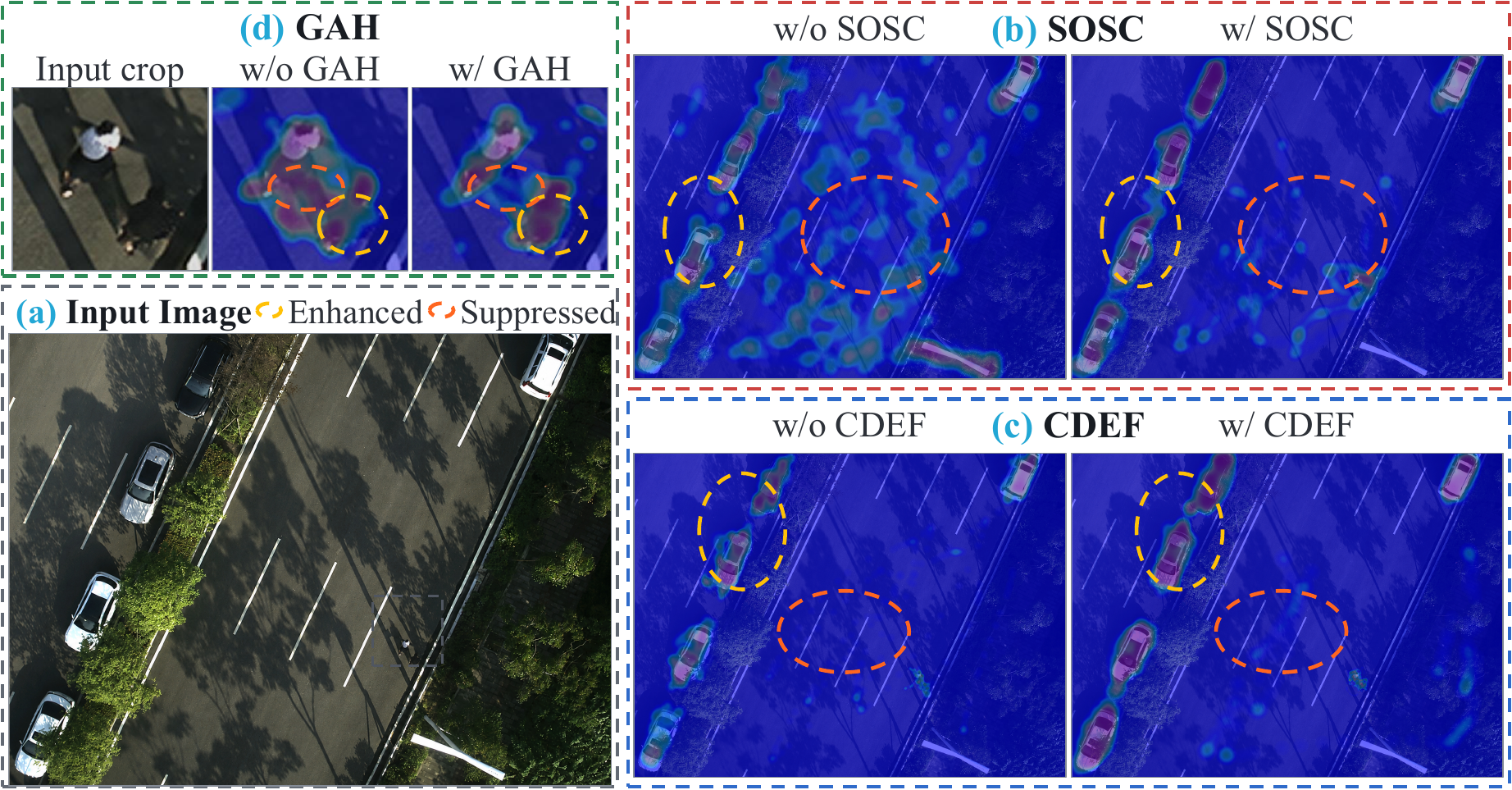}
\caption{\rev{Feature responses of S$^3$E-Det: (a) input image; (b)--(d) with/without SOSC, CDEF, and GAH. \textcolor{FeatYellow}{Yellow}/\textcolor{FeatOrange}{orange} dashed ellipses mark enhanced target/suppressed background responses, respectively.}}
\label{fig:component_feature_visualization}
\end{figure}

\rev{The with/without comparisons in Figure~\ref{fig:component_feature_visualization} provide a component-wise view of these response changes; yellow and orange dashed ellipses mark strengthened target responses and attenuated interference, respectively.} \rev{With SOSC, activations become more concentrated on vehicles and less dispersed over road texture.} \rev{With CDEF, the fused representation further consolidates target-related evidence while reducing responses in non-target regions.} \rev{GAH brings the local response extent into closer agreement with object geometry and limits mixed activations from adjacent areas.} \rev{Localized changes around small vehicles and residual dense-region activations indicate improved discrimination rather than complete background suppression.} \rev{The three stages show progressively more selective responses across spectral encoding, cross-level fusion, and geometric readout, consistent with the quantitative gains.}

\subsection{Ablation Studies and Analysis}

\noindent
To evaluate component contributions, design sensitivity, and rotation robustness, we conduct a series of ablation and analysis experiments on MODA. Unless otherwise specified, all studies follow the MODA input setting, training schedule, and evaluation protocol.

\noindent\textbf{Component Ablation.}
\revdel{} \rev{Table~\ref{tab:component_ablation} separates equivariance from the proposed modules.} \revdel{} \rev{The equivariant baseline gains 4.8\%, 6.1\%, and 4.4\% in mAP$_{50}$, mAP$_{75}$, and mAP.} \revdel{} \rev{On this baseline, SOSC, CDEF, and GAH individually improve mAP$_{50}$ by 2.8\%, 1.9\%, and 1.5\%, while full integration improves mAP$_{50}$, mAP$_{75}$, and mAP by 5.8\%, 5.6\%, and 6.3\%.} \revdel{} \rev{The full model yields the largest gains, confirming the complementary roles of SOSC, CDEF, and GAH.}

\begin{table}[!t]
\vspace{-4pt}
\centering
\scriptsize
\setlength{\tabcolsep}{3.2pt}
\caption{Ablation studies on the key components of S$^3$E-Det.}
\vspace{-4pt}
\label{tab:component_ablation}
\begin{tabular*}{\linewidth}{@{\extracolsep{\fill}}c|ccc|ccc@{}}
\toprule
Equiv. & SOSC & CDEF & GAH & mAP$_{50}\uparrow$ & mAP$_{75}\uparrow$ & mAP$\uparrow$ \\
\midrule
\xmark & \xmark & \xmark & \xmark & 62.0 & 48.0 & 43.3 \\
\cmark & \xmark & \xmark & \xmark & 66.8 & 54.1 & 47.7 \\
\cmark & \cmark & \xmark & \xmark & 69.6 & 55.5 & 50.0 \\
\cmark & \xmark & \cmark & \xmark & 68.7 & 54.9 & 49.4 \\
\cmark & \xmark & \xmark & \cmark & 68.3 & 55.2 & 49.2 \\
\cmark & \cmark & \cmark & \xmark & 71.3 & 58.3 & 52.3 \\
\cmark & \cmark & \xmark & \cmark & 71.2 & 58.0 & 52.0 \\
\cmark & \xmark & \cmark & \cmark & 70.0 & 56.5 & 50.8 \\
\cmark & \cmark & \cmark & \cmark & 72.6 & 59.7 & 54.0 \\
\bottomrule
\end{tabular*}
\vspace{-10pt}
\end{table}

\noindent\textbf{Hyperparameter Study on SOSC.}
Table~\ref{tab:sosc_hyperparams} evaluates the spectral kernel size and the two calibration gates in SOSC. A kernel size of 3 gives the best overall result, indicating that local neighboring bands provide sufficient context for intra-cell prediction. A kernel size of 1 lacks adjacent spectral evidence, whereas larger kernels introduce more distant bands with limited return. Removing either the residual gate \(\gamma_i\) or the derivative gate \(\eta_i\) reduces performance, showing that interpolated spectral correction and first-order variation modeling contribute jointly to order-aware calibration.

\begin{table}[!t]
\vspace{-4pt}
\centering
\scriptsize
\setlength{\tabcolsep}{2.6pt}
\caption{Hyperparameter ablation of SOSC on MODA.}
\vspace{-4pt}
\label{tab:sosc_hyperparams}
\begin{tabularx}{\linewidth}{@{}CCCCC@{}}
\toprule
\shortstack{Spectral\\kernel} & \shortstack{Residual\\gate \(\gamma\)} & \shortstack{Derivative\\gate \(\eta\)} & mAP$_{50}\uparrow$ & mAP$\uparrow$ \\
\midrule
1 & \cmark & \cmark & 71.3 & 52.6 \\
\rowcolor{AblationBestBlue}
3 & \cmark & \cmark & 72.6 & 54.0 \\
5 & \cmark & \cmark & 72.4 & 53.8 \\
7 & \cmark & \cmark & 72.2 & 53.5 \\
3 & \xmark & \cmark & 71.4 & 52.8 \\
3 & \cmark & \xmark & 71.0 & 52.3 \\
\bottomrule
\end{tabularx}
\vspace{-10pt}
\end{table}

\noindent\textbf{Hyperparameter Study on CDEF.}
Table~\ref{tab:cdef_hyperparams} analyzes the bottleneck reduction ratio, the spatial kernel size, and the agreement gate in CDEF. A reduction ratio of 4 with a \(3\times3\) spatial kernel yields the best accuracy. A smaller reduction ratio brings little additional benefit, while a larger ratio weakens the fused representation. Removing the agreement gate also decreases performance, indicating that agreement-guided fusion is useful for balancing common object context and differential spectral evidence.

\begin{table}[!t]
\vspace{-4pt}
\centering
\scriptsize
\setlength{\tabcolsep}{2.8pt}
\caption{Hyperparameter ablation of CDEF on MODA.}
\vspace{-4pt}
\label{tab:cdef_hyperparams}
\begin{tabularx}{\linewidth}{@{}CCCCC@{}}
\toprule
\shortstack{Reduction\\ratio} & \shortstack{Spatial\\kernel} & \shortstack{Agreement\\gate} & mAP$_{50}\uparrow$ & mAP$\uparrow$ \\
\midrule
2 & 3 & \cmark & 72.3 & 53.7 \\
\rowcolor{AblationBestBlue}
4 & 3 & \cmark & 72.6 & 54.0 \\
8 & 3 & \cmark & 71.9 & 53.3 \\
4 & 5 & \cmark & 72.4 & 53.8 \\
4 & 3 & \xmark & 71.6 & 52.9 \\
\bottomrule
\end{tabularx}
\vspace{-10pt}
\end{table}

\noindent\textbf{Rotation Group Order.}
Table~\ref{tab:group_order_ablation} studies how the discrete rotation group affects the accuracy--compute trade-off. Moving from \(C_1\) to \(C_2\) brings a clear gain, showing the benefit of explicit rotation modeling. \(C_4\) further improves the results and provides a favorable balance between orientation resolution and computation. Although \(C_8\) gives a marginal mAP$_{50}$ increase, it roughly doubles the parameters and FLOPs, making \(C_4\) the default choice.

\begin{table}[!t]
\vspace{-4pt}
\centering
\scriptsize
\setlength{\tabcolsep}{2.4pt}
\caption{Ablation of rotation group order on MODA.}
\vspace{-4pt}
\label{tab:group_order_ablation}
\begin{tabularx}{\linewidth}{@{}CCCCCC@{}}
\toprule
\shortstack{Group\\order} & mAP$_{50}\uparrow$ & mAP$_{75}\uparrow$ & mAP$\uparrow$ & Params$\downarrow$ & FLOPs$\downarrow$ \\
\midrule
\(C_1\) & 62.0 & 48.0 & 43.3 & 0.6M & 8.5G \\
\(C_2\) & 70.8 & 57.9 & 51.5 & 1.2M & 17.3G \\
\rowcolor{AblationBestBlue}
\(C_4\) & 72.6 & 59.7 & 54.0 & 2.4M & 35.6G \\
\(C_8\) & 72.7 & 59.8 & 54.0 & 5.0M & 73.0G \\
\bottomrule
\end{tabularx}
\vspace{-10pt}
\end{table}

\begin{figure*}[!t]
\begin{minipage}[t]{0.49\textwidth}
\centering
\includegraphics[width=\linewidth]{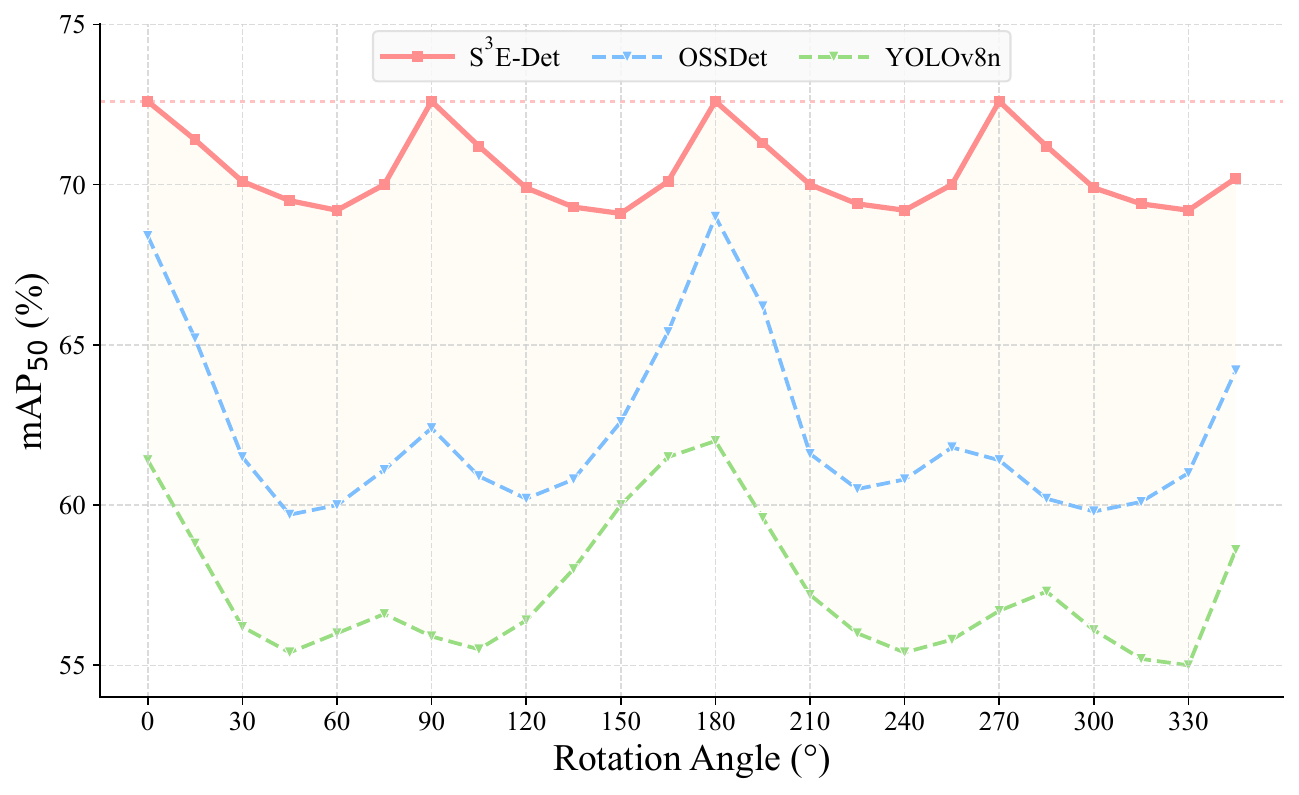}
\caption{Detection robustness under input rotations on MODA. mAP$_{50}$ is evaluated after rotating validation images from \(0^\circ\) to \(345^\circ\) in \(15^\circ\) steps.}
\label{fig:rotation_robustness}
\end{minipage}\hfill
\begin{minipage}[t]{0.49\textwidth}
\centering
\includegraphics[width=\linewidth]{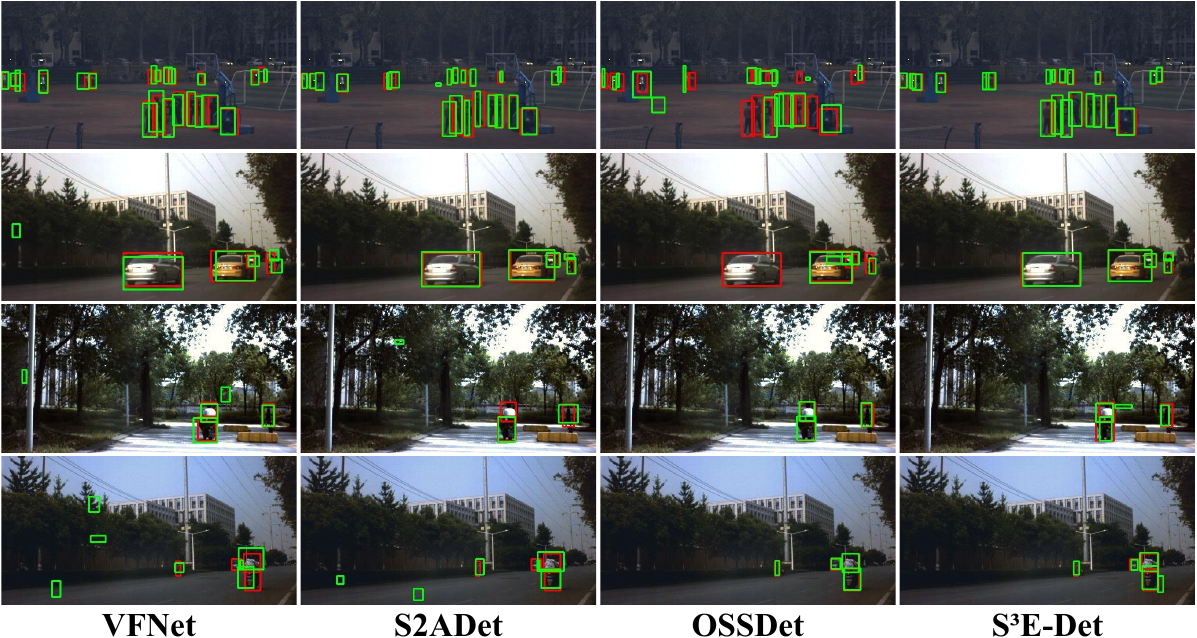}
\caption{Detection result comparison on HOD3K. \textcolor{red}{Red} boxes indicate the ground truth, while \textcolor{green}{green} boxes denote the predictions.}
\label{fig:hod3k_qualitative}
\end{minipage}
\end{figure*}

\begin{table}[!t]
\centering
\scriptsize
\setlength{\tabcolsep}{2.2pt}
\caption{\revdel{}\revsep\rev{Layer-wise cost and exact-$C_4$ residuals (mean over three non-identity elements). Means are computed from unrounded residuals; all displayed values lie in $[0,1]$.}}
\label{tab:eq_pipeline}
\begin{tabular*}{\columnwidth}{@{\extracolsep{\fill}}lrrrr@{}}
\toprule
Operator & \shortstack{Params $\downarrow$\\(2,445,776)} & \shortstack{FLOPs $\downarrow$\\(35.60G)} & \shortstack{Cumulative\\$\bar e_{C_4}\downarrow$} & \shortstack{Local\\$\bar c_{C_4}\downarrow$} \\
\midrule
L0 Stem (group lift) & 2,080 & 0.54G & 0.000000 & 0.000000 \\
L1 SOSC-1 & 10,307 & 2.59G & 0.000023 & 0.000004 \\
L2 Patch merging-1 & 2,112 & 0.34G & 0.000034 & 0.000004 \\
L3 SOSC-2 & 38,883 & 2.51G & 0.000227 & 0.000004 \\
L4 Patch merging-2 & 8,320 & 0.34G & 0.000059 & 0.000001 \\
L5 SOSC-3 & 151,331 & 2.46G & 0.000509 & 0.000005 \\
L6 Patch merging-3 & 33,024 & 0.34G & 0.000056 & 0.000008 \\
L7 SOSC-4 & 593,500 & 2.44G & 0.000100 & 0.000005 \\
L8 RE-SPPF & 65,792 & 0.27G & 0.001804 & 0.000005 \\
L9 FPN upsample-1 & 32,768 & 0.13G & 0.002022 & 0.000000 \\
L11 CDEF-1 & 97,825 & 1.42G & 0.001432 & 0.000031 \\
L12 FPN upsample-2 & 32,768 & 0.54G & 0.002040 & 0.000000 \\
L14 CDEF-2 (P3) & 56,353 & 3.27G & 0.000926 & 0.000025 \\
L15 FPN downsample-1 & 25,088 & 0.41G & 0.001150 & 0.000005 \\
L17 CDEF-3 (P4) & 223,297 & 3.25G & 0.001185 & 0.000003 \\
L18 FPN downsample-2 & 99,328 & 0.41G & 0.000905 & 0.000001 \\
L20 CDEF-4 (P5) & 389,500 & 1.43G & 0.000929 & 0.000001 \\
\addlinespace[1pt]
GAH-Classification readout & 324,800 & 7.37G & 0.000242 & 0.000001 \\
GAH-Canonical DFL readout & 201,200 & 4.29G & 0.001276 & 0.000023 \\
GAH-Phase-angle readout & 57,500 & 1.25G & 0.001233 & 0.000000 \\
Rotated-box decoder & -- & -- & 0.025145 & 0.022454 \\
Confidence filtering and NMS & -- & -- & 0.023452 & 0.000000 \\
\bottomrule
\end{tabular*}
\end{table}

\noindent\textbf{\rev{Exact-$C_4$ Implementation-Conformance Protocol.}}
\rev{Rotation accuracy measures robustness but does not verify the defining commutation identity.} \rev{Following layer-wise audits in MessDet~\addedcite{wu2025messdet} and the exact/inexact distinction on sampled grids~\addedcite{edixhoven2023}, we compute checkpoint-wise accumulated and operator-local residuals in FP32 on one fixed validation input.} \rev{Together with interpolation and prediction-set errors, these residuals distinguish deviation propagated through the feature hierarchy from that introduced by individual operators and decoding.} The checkpoint residual is
\begin{equation}
 e_l(\theta)=
 \frac{\left\|V_{l,\theta}\odot\left(\Phi_l(\mathcal{T}_{\theta}x)-
 \widetilde{\rho}_l(\theta)\Phi_l(x)\right)\right\|_2}
 {\left\|V_{l,\theta}\odot\widetilde{\rho}_l(\theta)\Phi_l(x)\right\|_2+\epsilon}
 ,
 \label{eq:layer_equivariance_error}
\end{equation}
where $l\in\mathcal{L}$ indexes the audited checkpoints, $V_{l,\theta}\in\{0,1\}^{H_l\times W_l}$ is the valid-support mask, and $\epsilon>0$ stabilizes the denominator. For an operator $m\in\mathcal{O}$ with input $Z$, the local commutator residual is
\begin{equation}
 c_m(\theta)=
 \frac{\left\|V_{m,\theta}\odot\left(m(\widetilde{\rho}_{\rm in}(\theta)Z)-
 \widetilde{\rho}_{\rm out}(\theta)m(Z)\right)\right\|_2}
 {\left\|V_{m,\theta}\odot\widetilde{\rho}_{\rm out}(\theta)m(Z)\right\|_2+\epsilon}
 .
 \label{eq:local_commutator_error}
\end{equation}
Here $V_{m,\theta}$ is the corresponding output-support mask.

\rev{For $\theta\in\{90^\circ,180^\circ,270^\circ\}$, inputs and features are transformed according to the discrete $C_4$ group action: spatial positions undergo the corresponding quarter turn, while orientation channels are cyclically permuted without interpolation.} \rev{We summarize exact-group conformance by $e^{C_4}_{\max}=\max_{l,\theta}e_l(\theta)$.} \rev{Intermediate angles use bilinear spatial rotation and fractional orientation shifts only to evaluate off-grid rotations, not as evidence of exact $C_4$ or continuous $\mathrm{SO}(2)$ equivariance~\addedcite{wu2025r2det}.}

\rev{We quantify the interpolation floor by the valid-support round-trip error}
\begin{equation}
 e_{\rm int}(\theta)=
 \frac{\left\|V_{\theta}\odot\left(\mathcal{T}_{-\theta}\mathcal{T}_{\theta}x-x\right)\right\|_2}
 {\left\|V_{\theta}\odot x\right\|_2+\epsilon}
 .
 \label{eq:interpolation_error}
\end{equation}
\rev{Here $V_{\theta}$ is the image-space valid-support mask and is propagated to every feature scale by nearest-neighbor resizing.} \rev{This control is important because interpolation and subsampling are sources of sampling-related equivariance error~\addedcite{edixhoven2023,zhang2019shift}.}

\rev{GAH outputs are evaluated under their response-specific laws.} \rev{Classification and canonical DFL use Eq.~\eqref{eq:layer_equivariance_error}; orientation uses $e_{\rm ang}=d_{\pi}(\hat\theta(\mathcal{T}_{\theta}x),\hat\theta(x)+\theta)/(\pi/2)$, where $d_{\pi}$ is angular distance modulo $\pi$, and $e_{\rm raw}=(e_{\rm cls}+e_{\rm DFL}+e_{\rm ang})/3$.} \rev{For decoded prediction sets $\mathcal{P}_{\theta}=\{(b_i,y_i)\}$ and $\mathcal{P}_{0}=\{(b_j,y_j)\}$, let $\mathcal{M}_{\theta}\subseteq\mathcal{P}_{\theta}\times\mathcal{P}_{0}$ be their ProbIoU-based class-aware Hungarian matching. Their normalized discrepancy is}
\begin{equation}
 e_{\rm set}(\theta)=1-
 \frac{2\!\sum_{(i,j)\in\mathcal{M}_{\theta}}
 \mathbf{1}[y_i=y_j]\operatorname{IoU}_{\rm r}(b_i,\mathcal{T}_{\theta}b_j)}
 {|\mathcal{P}_{\theta}|+|\mathcal{P}_{0}|},
 \label{eq:prediction_set_error}
\end{equation}
\rev{Unmatched and class-mismatched predictions are penalized.} \rev{All residuals are dimensionless and non-additive; displayed zeros denote values below reporting precision.}

\noindent\textbf{\rev{Exact-$C_4$ Implementation Conformance.}}
\rev{Tables~\ref{tab:eq_pipeline} and~\ref{tab:eq_c4_angles} localize numerical deviation by operator and rotation angle, respectively.} \rev{For exact quarter turns, the worst checkpoint residual is 0.002053, while the classification, canonical-DFL, and phase-angle residuals remain at or below 0.001276.} \rev{Patch merging, pointwise nonlinearities, and FPN resampling contribute only $10^{-6}$-scale local residuals.} \rev{The set discrepancy rises to 0.025145 after rotated-box decoding and remains similar after NMS (0.023452), placing the dominant implementation deviation in decoding rather than subsequent confidence filtering and NMS.} \rev{Together, these results show that, for the audited sample, the feature pathway numerically follows the discrete group action, while the three GAH branches provide invariant classification and canonical-DFL responses together with an equivariant orientation response.} \rev{This conclusion concerns implementation conformance on the fixed sample, not the error distribution over validation inputs.}

\noindent\textbf{\rev{Off-Grid Rotation Evaluation and Error Localization.}}
\rev{Table~\ref{tab:eq_c4_angles} also reveals a distinct error regime outside the discrete group: the mean interpolation residual increases to 0.013508, while the raw-head and post-NMS discrepancies reach 0.346815 and 0.662675.} \rev{These values quantify sensitivity to rotations that require interpolation and fractional orientation shifts; they neither contradict exact $C_4$ conformance nor establish continuous $\mathrm{SO}(2)$ equivariance.} \rev{Accordingly, the strict claim covers architecture-level commutation and response-specific raw predictions under exact quarter turns on compatible grids; off-grid rotations, confidence filtering, and NMS remain outside its scope.}

\noindent\textbf{Robustness to Input Rotations.}
To assess geometric consistency, we rotate MODA validation images by fixed angles. As shown in Figure~\ref{fig:rotation_robustness}, S$^3$E-Det shows smaller mAP$_{50}$ fluctuations and recovers at the \(C_4\) orthogonal angles, whereas OSSDet and YOLOv8n degrade more at intermediate angles. \revdel{}\revsep\rev{The periodic recovery is consistent with the discrete group structure used by the features and GAH.}

\begin{table*}[!t]
\centering
\scriptsize
\setlength{\tabcolsep}{3.5pt}
\caption{\revdel{}\revsep\rev{Angle-resolved implementation conformance on a $30^\circ$ grid. Quarter turns are exact $C_4$ elements; the remaining angles evaluate off-grid rotations. $e_{\rm raw}$ averages classification, canonical DFL, and phase-angle residuals. Mean rows are computed from unrounded residuals; all displayed values lie in $[0,1]$.}}
\label{tab:eq_c4_angles}
\begin{tabularx}{\textwidth}{@{}cCCCCCC@{}}
\toprule
Angle & $e_{\rm int}\downarrow$ & L8 backbone $\downarrow$ & \shortstack{P3--P5\\mean $\downarrow$} & $e_{\rm raw}\downarrow$ & \shortstack{Decoded\\OBBs $\downarrow$} & \shortstack{Post-NMS\\predictions $\downarrow$} \\
\midrule
$30^\circ$  & 0.013557 & 0.456471 & 0.376044 & 0.348256 & 0.432972 & 0.658733 \\
$60^\circ$  & 0.013669 & 0.547496 & 0.396743 & 0.359207 & 0.456248 & 0.663884 \\
$90^\circ$  & 0.000000 & 0.001776 & 0.001008 & 0.000890 & 0.022366 & 0.024853 \\
$120^\circ$ & 0.013473 & 0.467767 & 0.381041 & 0.336827 & 0.503154 & 0.768568 \\
$150^\circ$ & 0.013331 & 0.533103 & 0.383906 & 0.342756 & 0.451895 & 0.575583 \\
$180^\circ$ & 0.000000 & 0.001838 & 0.001013 & 0.000934 & 0.034079 & 0.033145 \\
$210^\circ$ & 0.013557 & 0.458926 & 0.374532 & 0.348194 & 0.433514 & 0.659853 \\
$240^\circ$ & 0.013669 & 0.545102 & 0.395484 & 0.359601 & 0.459182 & 0.671875 \\
$270^\circ$ & 0.000000 & 0.001796 & 0.001019 & 0.000927 & 0.018990 & 0.012359 \\
$300^\circ$ & 0.013473 & 0.466919 & 0.380692 & 0.336736 & 0.501414 & 0.771625 \\
$330^\circ$ & 0.013331 & 0.532164 & 0.384026 & 0.342943 & 0.451919 & 0.531280 \\
\midrule
Exact $C_4$ mean & 0.000000 & 0.001804 & 0.001013 & 0.000917 & 0.025145 & 0.023452 \\
Off-grid mean & 0.013508 & 0.500993 & 0.384059 & 0.346815 & 0.461287 & 0.662675 \\
\bottomrule
\end{tabularx}
\end{table*}

\subsection{Results on the HOD3K Dataset}

\noindent\textbf{Data and Setups.}
HOD3K~\shiftedcite{he2023s2adet} contains 3,242 \(512\times256\) images with 16 bands from 470 to 620 nm and 15,149 instances in people, cars, and bikes. It follows a split of 2,308 training, 219 validation, and 715 testing images. We use all 16 bands, axis-aligned IoU, an input resolution of \(1024\times512\), 14 training epochs, and a total batch size of 8.

\noindent\textbf{Quantitative Results.}
\revdel{} \rev{Table~\ref{tab:hod3k_comparison} evaluates the same S$^3$E-Det design under the 16-band HOD3K configuration, which is distinct from the eight-band MODA setting.} \revdel{} \rev{Relative to the published OSSDet results, the standard-augmentation row differs by 1.1\%, 1.0\%, and 0.7\% in mAP$_{50}$, mAP$_{75}$, and mAP, with 93.5\% fewer parameters and 79.0\% fewer FLOPs. Relative to the no-augmentation setting, standard augmentation improves these metrics by 0.8\%, 0.7\%, and 0.6\% without changing the model footprint. This isolates the contribution of augmentation within S$^3$E-Det.} \revdel{} \rev{Together, the results show that the accuracy--compactness pattern observed on MODA is retained under the 16-band input.}

\begin{table}[!t]
\vspace{-4pt}
\centering
\scriptsize
\setlength{\tabcolsep}{2.2pt}
\caption{Comparison with state-of-the-art detectors on HOD3K.}
\vspace{-4pt}
\label{tab:hod3k_comparison}
\begin{tabular*}{\linewidth}{@{\extracolsep{\fill}}lccccc@{}}
\toprule
Method & mAP$_{50}\uparrow$ & mAP$_{75}\uparrow$ & mAP$\uparrow$ & FLOPs$\downarrow$ & Params$\downarrow$ \\
\midrule
CO-DETR & 77.9 & 57.3 & 51.7 & 200.7G & 64.5M \\
DINO & 89.1 & 61.6 & 56.3 & 114.7G & 47.6M \\
Faster R-CNN & 89.4 & 62.7 & 56.9 & 206.7G & 41.1M \\
ATSS & 91.2 & 63.0 & 57.4 & 110.6G & 32.2M \\
TOOD & 90.3 & 65.6 & 57.8 & 108.5G & 32.1M \\
RepPoints & 92.0 & 65.5 & 58.4 & 104.5G & 36.9M \\
VFNet & 92.3 & 66.5 & 59.0 & 104.5G & 32.8M \\
C-RCNN & 90.9 & 66.9 & 59.6 & 149.5G & 69.2M \\
S$^2$ADet~\shiftedcite{he2023s2adet} & 93.4 & -- & 59.8 & 169.2G & 48.6M \\
OSSDet~\shiftedcite{han2025moda} & 93.4 & 68.8 & 60.9 & 136.2G & 37.7M \\
\midrule
S$^3$E-Det (no aug.) & 93.7 & 69.1 & 61.0 & \textbf{28.6G} & \textbf{2.4M} \\
\rev{S$^3$E-Det (standard aug.)} & \rev{\textbf{94.5}} & \rev{\textbf{69.8}} & \rev{\textbf{61.6}} & \rev{\textbf{28.6G}} & \rev{\textbf{2.4M}} \\
\bottomrule
\end{tabular*}
\vspace{-10pt}
\end{table}

\noindent\textbf{Qualitative Results.}
As shown in Figure~\ref{fig:hod3k_qualitative}, competing detectors tend to miss small or partially occluded instances that blend into roads and buildings, and they introduce false positives in low-light clutter. Across the shown examples, S$^3$E-Det exhibits fewer missed detections and spurious responses. \revdel{}\revsep\rev{Overall, the examples support more consistent detection across road, vegetation, and building backgrounds, while the remaining errors are dominated by small, occluded, or low-visibility instances.}

\Needspace{4\baselineskip}
\subsection{Results on the DroneVehicle Dataset}

\noindent\textbf{Data and Setups.}
DroneVehicle~\shiftedcite{dronevehicle} contains 28,439 visible--infrared pairs and 953,087 oriented annotations in five vehicle categories, split into 17,990/1,469/8,980 train/validation/test pairs. We concatenate each pair as a four-channel input and train at \(640\times512\) for 12 epochs with batch size 16; other settings follow MODA.

\noindent\textbf{Quantitative Results.}
\revdel{} \rev{Table~\ref{tab:dronevehicle_comparison} complements the many-band experiments with paired RGB--infrared detection.} \revdel{} \rev{Relative to the published WaveMamba results, the standard-augmentation row differs by 2.0\% in mAP$_{50}$ and 1.6\% in mAP; the corresponding differences from DMM are 4.6\% and 6.3\%.} \revdel{} \rev{Unlike the RGB--infrared-specific baselines, S$^3$E-Det retains the architecture evaluated on MODA and HOD3K.} \rev{The same architecture remains competitive across the evaluated multispectral configurations and attains the best result in four of five DroneVehicle categories.} \rev{Relative to no augmentation, standard augmentation improves the car, bus, truck, freight-car, and van scores by 1.3\%, 0.7\%, 1.2\%, 0.4\%, and 0.9\%, respectively.} \rev{Both S$^3$E-Det variants retain 2.4M parameters and require 10.9G FLOPs under the four-channel input.} \rev{Together, these results extend the observed accuracy--compactness pattern from many-band inputs to paired visible--infrared detection.}

\begin{figure}[!t]
\centering
\includegraphics[width=0.95\columnwidth]{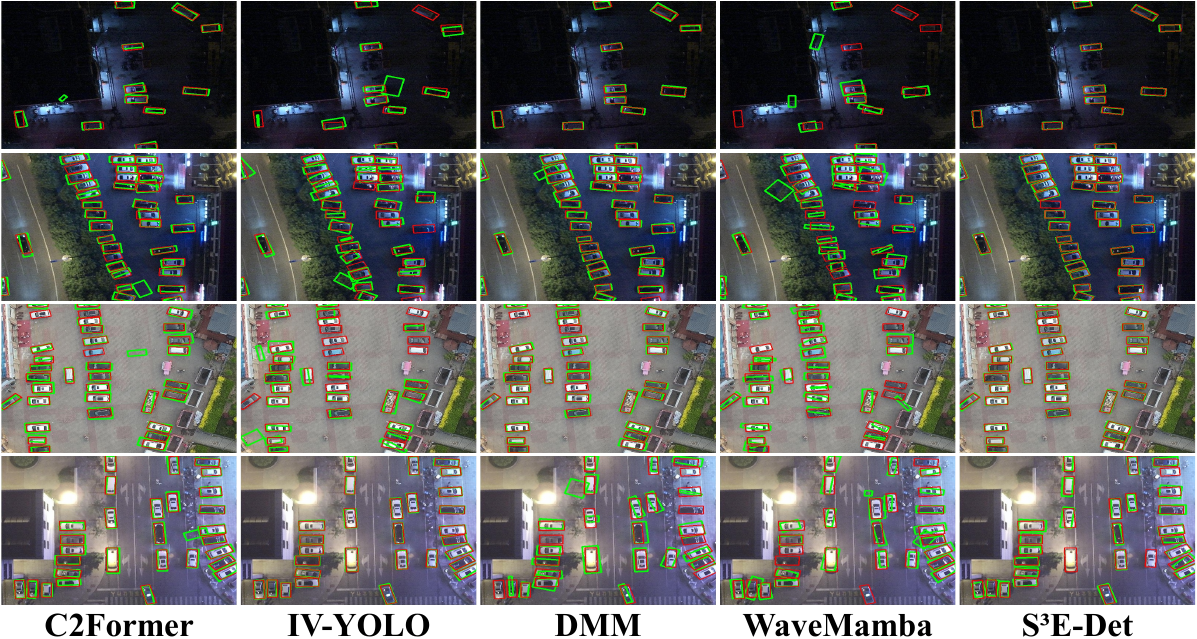}
\caption{Detection result comparison on DroneVehicle. \textcolor{red}{Red} boxes indicate the ground truth, while \textcolor{green}{green} boxes denote the predictions.}
\label{fig:dronevehicle_qualitative}
\end{figure}

\begin{table}[!t]
\vspace{-4pt}
\centering
\scriptsize
\setlength{\tabcolsep}{1.4pt}
\caption{Performance comparison on DroneVehicle.\revdel{}}
\vspace{-4pt}
\label{tab:dronevehicle_comparison}
\begin{tabular*}{\linewidth}{@{\extracolsep{\fill}}lccccccc@{}}
\toprule
Method & Car & Bus & Truck & Fre. & Van & mAP$_{50}\uparrow$ & mAP$\uparrow$ \\
\midrule
C$^2$Former~\shiftedcite{c2former} & 90.2 & 89.8 & 68.3 & 64.4 & 58.5 & 74.2 & 47.3 \\
YOLOv8l-RGB~\shiftedcite{yolov8} & 92.5 & 91.7 & 68.8 & 47.8 & 50.2 & 70.2 & 48.6 \\
YOLOv8l-IR~\shiftedcite{yolov8} & 93.4 & 91.9 & 69.3 & 53.7 & 51.1 & 71.9 & 49.1 \\
SLBAF-Net~\shiftedcite{slbaf_net} & 90.2 & 89.9 & 76.0 & 68.2 & 59.9 & 76.8 & 49.5 \\
CSOM-ODAF~\shiftedcite{csom_odaf} & 90.1 & 89.8 & 75.6 & 68.2 & 61.8 & 77.1 & 50.1 \\
CMA~\shiftedcite{m2fnet} & 95.8 & 93.1 & 75.9 & 59.8 & 59.4 & 76.8 & 50.4 \\
GLFNet~\shiftedcite{glfnet} & 90.3 & 88.0 & 72.7 & 53.6 & 52.6 & 71.4 & 54.8 \\
DMM~\shiftedcite{dmm} & 90.4 & 88.7 & 77.8 & 63.0 & 66.0 & 77.2 & 55.8 \\
IV-YOLO~\shiftedcite{iv_yolo} & 97.2 & \textbf{94.3} & 65.4 & 63.1 & 53.0 & 74.6 & 56.8 \\
WaveMamba~\shiftedcite{wavemamba} & 95.0 & 90.6 & 80.4 & 68.5 & 64.5 & 79.8 & 60.5 \\
\midrule
S$^3$E-Det (no aug.) & 96.5 & 93.0 & 79.5 & 68.5 & 67.0 & 80.9 & 61.5 \\
\rev{S$^3$E-Det (standard aug.)} & \rev{\textbf{97.8}} & \rev{93.7} & \rev{\textbf{80.7}} & \rev{\textbf{68.9}} & \rev{\textbf{67.9}} & \rev{\textbf{81.8}} & \rev{\textbf{62.1}} \\
\bottomrule
\end{tabular*}
\vspace{-10pt}
\end{table}
\Needspace{5\baselineskip}
\noindent\textbf{Qualitative Results.}
Figure~\ref{fig:dronevehicle_qualitative} shows that competing methods miss small, overlapping, or illumination-inconsistent vehicles and produce false positives in dense traffic. \revdel{}\revsep\rev{S$^3$E-Det shows fewer misses and background false positives, with predictions better aligned with the annotations.} \rev{Overall, the examples support more consistent detection under illumination and density variation, while the remaining errors are concentrated around small, occluded, or adjacent vehicles whose visible boundaries overlap. Across the displayed traffic layouts, this tendency persists from sparse, well-lit roads to crowded or illumination-inconsistent scenes with closely spaced vehicles and partially overlapping boundaries.}

\FloatBarrier
\Needspace{8\baselineskip}
\subsection{\rev{On-Device Inference Evaluation}}

\rev{Group-order and component variants are converted to FP16 RKNN and evaluated on an Orange Pi 5 Plus (RK3588) using RKNN Runtime 2.3.2, RKNPU driver 0.9.6, three 1.0-GHz NPU cores, batch size 1, and a padded $1216\times928$ input.} \rev{Statistics use 30 runs after 10 warm-ups; timing covers RKNN execution and output retrieval while excluding input loading, preprocessing, rotated-box decoding, and NMS.} \rev{Peak memory denotes the host process's maximum resident set size, average board power is measured at the 12~V DC input, and energy per image is the product of mean latency and average power.}

\rev{Table~\ref{tab:edge_deployment} exposes two distinct cost trends. Across rotation-group variants, latency and peak memory grow with group order: from $C_1$ to $C_8$, latency rises from 28.2 to 181.0~ms, throughput falls from 35.5 to 5.5~FPS, and peak memory increases from 82 to 595~MiB. Average board power grows more gradually, from 5.7 to 11.2~W, so the longer execution raises energy from 0.16 to 2.03~J/image. Within the $C_4$ component sweep, SOSC is the dominant incremental cost, adding 27.0~ms and 105~MiB to the equivariant baseline; CDEF and GAH add 15.0--17.0~ms and 53--67~MiB. The complete model operates at 89.3~ms, 11.2~FPS, 312.4~MiB, and 0.82~J/image. These measurements identify group expansion and SOSC as the principal deployment bottlenecks, while CDEF and GAH retain smaller device-level footprints.}

\begin{table}[!htbp]
\centering
\scriptsize
\setlength{\tabcolsep}{1.0pt}
\renewcommand{\arraystretch}{1.18}
\caption{\rev{Measured FP16 RKNN deployment costs of rotation-group and component variants.}}
\label{tab:edge_deployment}
{%
\ifmarkedchanges\color{blue}\fi
\begin{tabular*}{\columnwidth}{@{\extracolsep{\fill}}lccccc@{}}
\toprule
Configuration & \shortstack{Mean\\(ms/image)$\downarrow$} & FPS$\uparrow$ &
\shortstack{Peak\\Memory (MiB)$\downarrow$} & \shortstack{Avg. Board\\Power (W)$\downarrow$} & \shortstack{Energy\\(J/image)$\downarrow$} \\
\midrule
\multicolumn{6}{c}{\textit{Rotation-group variants}} \\
$C_1$ & 28.2 & 35.5 & 82 & 5.7 & 0.16 \\
$C_2$ & 56.0 & 17.9 & 168 & 8.0 & 0.45 \\
$C_4$ & 89.3 & 11.2 & 312.4 & 9.2 & 0.82 \\
$C_8$ & 181.0 & 5.5 & 595 & 11.2 & 2.03 \\
\midrule
\multicolumn{6}{c}{\textit{Component variants at $C_4$}} \\
Equivariant baseline & 57.0 & 17.5 & 175 & 7.5 & 0.43 \\
$+$SOSC & 84.0 & 11.9 & 280 & 8.6 & 0.72 \\
$+$CDEF & 72.0 & 13.9 & 242 & 8.1 & 0.58 \\
$+$GAH & 74.0 & 13.5 & 228 & 8.4 & 0.62 \\
Full S$^3$E-Det & 89.3 & 11.2 & 312.4 & 9.2 & 0.82 \\
\bottomrule
\end{tabular*}
}
\end{table}
\Needspace{4\baselineskip}
\section{Conclusions}

We present S$^3$E-Det, a strict spatial--spectral equivariant detector for MOD.
\revdel{} \rev{Unlike fusion-centered MOD detectors, S$^3$E-Det unifies order-preserving spectral calibration, scale-adaptive equivariant fusion, and response-specific geometric decoding.} \revdel{} \rev{Experiments on MODA, HOD3K, and DroneVehicle establish a favorable accuracy--compactness trade-off and consistent rotation robustness, while ablations isolate the complementary contributions of SOSC, CDEF, and GAH.}
\rev{Separate training within each benchmark leaves zero-shot cross-sensor transfer, controlled sensor perturbations, and large-corpus scaling unresolved.} \rev{Future work will extend evaluation across sensors, imaging conditions, and larger multispectral datasets, including wavelength-aware adaptation through spectral pretraining.}


\begin{thebibliography}{999}
%
\bibitem{vali2020deep}
Vali, A.; Comai, S.; Matteucci, M. Deep learning for land use and land cover classification based on hyperspectral and multispectral earth observation data: A review. {\em Remote Sens.} {\bf 2020}, {\em 12}, 2495.

%
\bibitem{dronevehicle}
Sun, Y.; Cao, B.; Zhu, P.; Hu, Q. Drone-based RGB-infrared cross-modality vehicle detection via uncertainty-aware learning. {\em IEEE Trans. Circuits Syst. Video Technol.} {\bf 2022}, {\em 32}, 6700--6713.

%
\bibitem{han2025moda}
Han, S.; Xu, T.; Liu, P.; Li, J. MODA: The first challenging benchmark for multispectral object detection in aerial images. In {\em Proceedings of the AAAI Conference on Artificial Intelligence}, Singapore, 20--27 January 2026; Volume 40, pp. 4574--4582.

%
%
\bibitem{sun2019hyperspectral}
Sun, W.; Du, Q. Hyperspectral band selection: A review. {\em IEEE Geosci. Remote Sens. Mag.} {\bf 2019}, {\em 7}, 118--139.

%
\bibitem{ma2025lightweight}
Ma, X.; Zhang, M.; Kan, Z.; Gao, Y.; Li, W. A lightweight spatial--spectral deformable CNN for UAV hyperspectral image classification. {\em IEEE Trans. Geosci. Remote Sens.} {\bf 2025}, {\em 63}, 5524213.

\bibitem{hong2026hyperspectral}
\rev{Hong, D.; Li, C.; Yokoya, N.; Zhang, B.; Jia, X.; Plaza, A.; Gamba, P.; Benediktsson, J.A.; Chanussot, J. Hyperspectral imaging.} {\em \rev{Nat. Rev. Methods Primers}} {\bf \rev{2026}}, {\em \rev{6}}, \rev{19.}

\bibitem{he2023s2adet}
He, X.; Tang, C.; Liu, X.; Zhang, W.; Sun, K.; Xu, J. Object detection in hyperspectral image via unified spectral--spatial feature aggregation. {\em IEEE Trans. Geosci. Remote Sens.} {\bf 2023}, {\em 61}, 5521213.

\bibitem{cai2022mst}
\rev{Cai, Y.; Lin, J.; Hu, X.; Wang, H.; Yuan, X.; Zhang, Y.; Timofte, R.; Van Gool, L. Mask-guided spectral-wise transformer for efficient hyperspectral image reconstruction. In }{\em \rev{Proceedings of the IEEE/CVF Conference on Computer Vision and Pattern Recognition (CVPR)}}\rev{, New Orleans, LA, USA, 19--24 June 2022; pp. 17502--17511.}

\bibitem{li2026im2state}
\rev{Li, C.; Hong, D.; Zhang, B.; Chanussot, J. Im2State: Image-to-state Mamba for hyperspectral and multispectral image fusion.} {\em \rev{Inf. Fusion}} {\bf \rev{2026}}, {\em \rev{134}}, \rev{104385.}

\bibitem{hong2024spectralgpt}
\rev{Hong, D.; Zhang, B.; Li, X.; Li, Y.; Li, C.; Yao, J.; Yokoya, N.; Li, H.; Ghamisi, P.; Jia, X.; Plaza, A.; Gamba, P.; Benediktsson, J.A.; Chanussot, J. SpectralGPT: Spectral remote sensing foundation model.} {\em \rev{IEEE Trans. Pattern Anal. Mach. Intell.}} {\bf \rev{2024}}, {\em \rev{46}}, \rev{5227--5244.}

%
\bibitem{dai2017dcn}
Dai, J.; Qi, H.; Xiong, Y.; Li, Y.; Zhang, G.; Hu, H.; Wei, Y. Deformable convolutional networks. In {\em Proceedings of the IEEE International Conference on Computer Vision}, Venice, Italy, 22--29 October 2017; pp. 764--773.

%
\bibitem{cohen2016gcnn}
Cohen, T.; Welling, M. Group equivariant convolutional networks. In {\em Proceedings of the 33rd International Conference on Machine Learning}; PMLR: New York, NY, USA, 2016; Volume 48, pp. 2990--2999.

%
\bibitem{han2021redet}
Han, J.; Ding, J.; Xue, N.; Xia, G.-S. ReDet: A rotation-equivariant detector for aerial object detection. In {\em Proceedings of the IEEE/CVF Conference on Computer Vision and Pattern Recognition}, Virtual, 19--25 June 2021; pp. 2786--2795.

%
\bibitem{wu2025messdet}
Wu, X.; Wang, X.; Zhu, X.; Yang, L.; Liu, J.; Hu, X. Measuring the impact of rotation equivariance on aerial object detection. In {\em Proceedings of the IEEE/CVF International Conference on Computer Vision}, Honolulu, HI, USA, 19--23 October 2025; pp. 7329--7339.

\bibitem{edixhoven2023}
\rev{Edixhoven, T.; Lengyel, A.; van Gemert, J.C. Using and abusing equivariance. In }{\em \rev{Proceedings of the IEEE/CVF International Conference on Computer Vision Workshops}}
\rev{, Paris, France, 2--6 October 2023; pp. 119--128.}

\bibitem{zhang2019shift}
\rev{Zhang, R. Making convolutional networks shift-invariant again. In }{\em \rev{Proceedings of the 36th International Conference on Machine Learning}}
\rev{, Long Beach, CA, USA, 9--15 June 2019; Volume 97, pp. 7324--7334.}

%
\bibitem{wu2025r2det}
Wu, Z.; Liu, Y.; Dong, H.; Tang, X.; Yang, J.; Jin, B.; Chen, M.; Wei, X. R2Det: Exploring relaxed rotation equivariance in 2D object detection. In {\em Proceedings of the International Conference on Learning Representations}, Singapore, 24--28 April 2025.

\bibitem{oriented_rcnn}
Xie, X.; Cheng, G.; Wang, J.; Yao, X.; Han, J. Oriented R-CNN for object detection. In {\em Proceedings of the IEEE/CVF International Conference on Computer Vision}, Montreal, QC, Canada, 11--17 October 2021; pp. 3520--3529.

\bibitem{lsknet}
Li, Y.; Hou, Q.; Zheng, Z.; Cheng, M.-M.; Yang, J.; Li, X. LSKNet: A foundation lightweight backbone for remote sensing. {\em Int. J. Comput. Vis.} {\bf 2024}, {\em 132}, 3736--3752.

\bibitem{strip_rcnn}
Yuan, X.; Zheng, Z.; Li, Y.; Liu, X.; Liu, L.; Li, X.; Hou, Q.; Cheng, M.-M. Strip R-CNN: Large strip convolution for remote sensing object detection. In {\em Proceedings of the AAAI Conference on Artificial Intelligence}, Singapore, 20--27 January 2026; Volume 40, pp. 12259--12267.

\bibitem{s2anet}
Han, J.; Ding, J.; Li, J.; Xia, G.-S. Align deep features for oriented object detection. {\em IEEE Trans. Geosci. Remote Sens.} {\bf 2022}, {\em 60}, 1--11.

\bibitem{r3det_kld}
Yang, X.; Yang, X.; Yang, J.; Ming, Q.; Wang, W.; Tian, Q.; Yan, J. Learning high-precision bounding box for rotated object detection via Kullback--Leibler divergence. In {\em Advances in Neural Information Processing Systems}; 2021; Volume 34, pp. 18381--18394.

\bibitem{oriented_reppoints}
Li, W.; Chen, Y.; Hu, K.; Zhu, J. Oriented RepPoints for aerial object detection. In {\em Proceedings of the IEEE/CVF Conference on Computer Vision and Pattern Recognition}, New Orleans, LA, USA, 18--24 June 2022; pp. 1829--1838.

\bibitem{dcfl}
Xu, C.; Ding, J.; Wang, J.; Yang, W.; Yu, H.; Yu, L.; Xia, G.-S. Dynamic coarse-to-fine learning for oriented tiny object detection. In {\em Proceedings of the IEEE/CVF Conference on Computer Vision and Pattern Recognition}, Vancouver, BC, Canada, 17--24 June 2023; pp. 7318--7328.

\bibitem{arsdetr}
Zeng, Y.; Chen, Y.; Yang, X.; Li, Q.; Yan, J. ARS-DETR: Aspect ratio-sensitive detection transformer for aerial oriented object detection. {\em IEEE Trans. Geosci. Remote Sens.} {\bf 2024}, {\em 62}, 1--15.

\bibitem{rhino}
\rev{Lee, H.; Song, M.; Koo, J.; Seo, J. Hausdorff distance matching with adaptive query denoising for rotated detection transformer. In }{\em \rev{Proceedings of the IEEE/CVF Winter Conference on Applications of Computer Vision (WACV)}}\rev{, Tucson, AZ, USA, 28 February--4 March 2025; pp. 1872--1882.}

\bibitem{ori_detr}
Zhao, Z.; Xue, Q.; He, Y.; Bai, Y.; Wei, X.; Gong, Y. Projecting points to axes: Oriented object detection via point-axis representation. In {\em Proceedings of the European Conference on Computer Vision}, Milan, Italy, 29 September--4 October 2024; pp. 161--179.

\bibitem{c2former}
Yuan, M.; Wei, X. C$^2$Former: Calibrated and complementary transformer for RGB-infrared object detection. {\em IEEE Trans. Geosci. Remote Sens.} {\bf 2024}, {\em 62}, 1--12.

\bibitem{yolov8}
Jocher, G.; Chaurasia, A.; Qiu, J. Ultralytics YOLOv8, version 8.0.0. 2023. Available online: \url{https://github.com/ultralytics/ultralytics}.

\bibitem{yolo11}
Jocher, G.; Qiu, J. Ultralytics YOLO11, version 11.0.0. 2024. Available online: \url{https://github.com/ultralytics/ultralytics}.

\bibitem{yolov12}
Tian, Y.; Ye, Q.; Doermann, D. YOLOv12: Attention-centric real-time object detectors. {\em arXiv} {\bf 2025}, arXiv:2502.12524.

\bibitem{slbaf_net}
Cheng, X.; Geng, K.; Wang, Z.; Wang, J.; Sun, Y.; Ding, P. SLBAF-Net: Super-lightweight bimodal adaptive fusion network for UAV detection in low recognition environment. {\em Multimed. Tools Appl.} {\bf 2023}, {\em 82}, 47773--47792.

\bibitem{csom_odaf}
Chen, C.; Qi, J.; Liu, X.; Bin, K.; Fu, R.; Hu, X.; Zhong, P. Weakly misalignment-free adaptive feature alignment for UAVs-based multimodal object detection. In {\em Proceedings of the IEEE/CVF Conference on Computer Vision and Pattern Recognition}, Seattle, WA, USA, 17--21 June 2024; pp. 26836--26845.

\bibitem{m2fnet}
Jiang, C.; Ren, H.; Yang, H.; Huo, H.; Zhu, P.; Yao, Z.; Li, J.; Sun, M.; Yang, S. M2FNet: Multi-modal fusion network for object detection from visible and thermal infrared images. {\em Int. J. Appl. Earth Obs. Geoinf.} {\bf 2024}, {\em 130}, 103918.

\bibitem{glfnet}
Kang, X.; Yin, H.; Duan, P. Global-local feature fusion network for visible-infrared vehicle detection. {\em IEEE Geosci. Remote Sens. Lett.} {\bf 2024}, {\em 21}, 1--5.

\bibitem{dmm}
Zhou, M.; Li, T.; Qiao, C.; Xie, D.; Wang, G.; Ruan, N.; Mei, L.; Yang, Y.; Tao Shen, H. DMM: Disparity-guided multispectral mamba for oriented object detection in remote sensing. {\em IEEE Trans. Geosci. Remote Sens.} {\bf 2025}, {\em 63}, 1--13.

\bibitem{iv_yolo}
Tian, D.; Yan, X.; Zhou, D.; Wang, C.; Zhang, W. IV-YOLO: A lightweight dual-branch object detection network. {\em Sensors} {\bf 2024}, {\em 24}, 6181.

\bibitem{wavemamba}
Zhu, H.; Dong, W.; Yang, L.; Li, H.; Yang, Y.; Ren, Y.; Zhu, Q.; Feng, Z.; Li, C.; Lin, S.; Wang, R.; Luo, X.; Zhang, B. WaveMamba: Wavelet-driven mamba fusion for RGB-infrared object detection. In {\em Proceedings of the IEEE/CVF International Conference on Computer Vision}, Honolulu, HI, USA, 19--23 October 2025; pp. 11219--11229.

\end{thebibliography}
\end{document}